\documentclass[journal]{IEEEtran}
\usepackage{cite}
\usepackage{hyperref}

\ifCLASSINFOpdf
  \usepackage[pdftex]{graphicx}
   \graphicspath{{../pdf/}{../jpeg/}}
   \DeclareGraphicsExtensions{.pdf,.jpeg,.png,.jpg}
\else
  \usepackage[dvips]{graphicx}
   \graphicspath{{../eps/}}
   \DeclareGraphicsExtensions{.eps}
\fi
\usepackage{amsmath}
\usepackage{algorithmic}
\usepackage{array}
\ifCLASSOPTIONcompsoc
  \usepackage[caption=false,font=normalsize,labelfont=sf,textfont=sf]{subfig}
\else
  \usepackage[caption=false,font=footnotesize]{subfig}
\fi
\usepackage{url}
\usepackage{multirow}
\usepackage{multicol}
\usepackage{booktabs}
\usepackage{chngpage}
\usepackage{threeparttable}
\usepackage{bbm}
\begin{document}
\title{Object Detection with Deep Learning: A Review}

\author{Zhong-Qiu Zhao,~\IEEEmembership{Member,~IEEE,}
        Peng Zheng,\\
        Shou-tao Xu,
        and~Xindong~Wu,~\IEEEmembership{~Fellow,~IEEE}% <-this % stops a space
\thanks{Zhong-Qiu Zhao, Peng Zheng and Shou-Tao Xu are with the College of Computer
Science and Information Engineering, Hefei University of Technology, China.
Xindong Wu is with the School of Computing and Informatics,
University of Louisiana at Lafayette, USA.}
%Zhong-Qiu Zhao is the corresponding author, E-mail: z.zhao@hfut.edu.cn.}% <-this % stops a space
\thanks{Manuscript received August xx, 2017; revised xx xx, 2017.}
}

\markboth{This paper has been accepted by  IEEE Transactions on Neural Networks and Learning Systems for publication}%
{Zhao \MakeLowercase{\textit{et al.}}: Object Detection with Deep Learning: A Review}



\maketitle

\begin{abstract}
Due to object detection's close relationship with video analysis and image understanding, it has attracted much research attention in recent years. Traditional object detection methods are built on handcrafted features and shallow trainable architectures. Their performance easily stagnates by constructing complex ensembles which combine multiple low-level image features with high-level context from object detectors and scene classifiers. With the rapid development in deep learning, more powerful tools, which are able to learn semantic, high-level, deeper features, are introduced to address the problems existing in traditional architectures. These models behave differently in network architecture, training strategy and optimization function, etc. In this paper, we provide a review on deep learning based object detection frameworks. Our review begins with a brief introduction on the history of deep learning and its representative tool, namely Convolutional Neural Network (CNN). Then we focus on typical generic object detection architectures along with some modifications and useful tricks to improve detection performance further. As distinct specific detection tasks exhibit different characteristics, we also briefly survey several specific tasks, including salient object detection, face detection and pedestrian detection. Experimental analyses are also provided to compare various methods and draw some meaningful conclusions. Finally, several promising directions and tasks are provided to serve as guidelines for future work in both object detection and relevant neural network based learning systems.
\end{abstract}


% \fi
\IEEEpeerreviewmaketitle

\vspace{-0.5cm}
\section{Introduction}
\label{sec:intro}
\IEEEPARstart{T}{o} gain a complete image understanding, we should not only concentrate on classifying different images, but also try to precisely estimate the concepts and locations of objects contained in each image. This task is referred as object detection \cite{Felzenszwalb2010Object}[S1], which usually consists of different subtasks such as face detection \cite{Sung2002Example}[S2], pedestrian detection \cite{Wojek2012Pedestrian}[S2] and skeleton detection \cite{Kobatake1996Detection}[S3]. As one of the fundamental computer vision problems, object detection is able to provide valuable information for semantic understanding of images and videos, and is related to many applications, including image classification \cite{jia2014caffe,alexnet}, human behavior analysis \cite{Cao2017RealtimeM2}[S4], face recognition \cite{yang2016multi}[S5] and autonomous driving \cite{Chen2015DeepDrivingLA,Chen2017Multi}. Meanwhile, Inheriting from neural networks and related learning systems, the progress in these fields will develop neural network algorithms, and will also have great impacts on object detection techniques which can be considered as learning systems.\cite{Dundar2017Embedded, Cintra2018Low, Khan2017Cost, Stuhlsatz2012Feature}[S6]. However, due to large variations in viewpoints, poses, occlusions and lighting conditions, it's difficult to perfectly accomplish object detection with an additional object localization task. So much attention has been attracted to this field in recent years \cite{rcnn,frcn,yolo, Faster}.

The problem definition of object detection is to determine where objects are located in a given image (object localization) and which category each object belongs to (object classification). So the pipeline of traditional object detection models can be mainly divided into three stages: informative region selection, feature extraction and classification.

\noindent\textbf{Informative region selection.} As different objects may appear in any positions of the image and have different aspect ratios or sizes, it is a natural choice to scan the whole image with a multi-scale sliding window. Although this exhaustive strategy can find out all possible positions of the objects, its shortcomings are also obvious. Due to a large number of candidate windows, it is computationally expensive and produces too many redundant windows. However, if only a fixed number of sliding window templates are applied, unsatisfactory regions may be produced.

\noindent\textbf{Feature extraction.} To recognize different objects, we need to extract visual features which can provide a semantic and robust representation. SIFT \cite{SIFT}, HOG \cite{HOG} and Haar-like \cite{haar-like} features are the representative ones. This is due to the fact that these features can produce representations associated with complex cells in human brain \cite{SIFT}. However, due to the diversity of appearances, illumination conditions and backgrounds, it's difficult to manually design a robust feature descriptor to perfectly describe all kinds of objects.

\noindent\textbf{Classification. }Besides, a classifier is needed to distinguish a target object from all the other categories and to make the representations more hierarchical, semantic and informative for visual recognition. Usually, the Supported Vector Machine (SVM) \cite{svm}, AdaBoost \cite{adaboost} and Deformable Part-based Model (DPM) \cite{DPM} are good choices. Among these classifiers, the DPM is a flexible model by combining object parts with deformation cost to handle severe deformations. In DPM, with the aid of a graphical model, carefully designed low-level features and kinematically inspired part decompositions are combined. And discriminative learning of graphical models allows for building high-precision part-based models for a variety of object classes.

Based on these discriminant local feature descriptors and shallow learnable architectures, state of the art results have been obtained on PASCAL VOC object detection competition \cite{voc2007} and real-time embedded systems have been obtained with a low burden on hardware. However, small gains are obtained during 2010-2012 by only building ensemble systems and employing minor variants of successful methods \cite{rcnn}. This fact is due to the following reasons: 1) The generation of candidate bounding boxes with a sliding window strategy is redundant, inefficient and inaccurate. 2) The semantic gap cannot be bridged by the combination of manually engineered low-level descriptors and discriminatively-trained shallow models.

Thanks to the emergency of Deep Neural Networks (DNNs) \cite{alexnet}[S7], a more significant gain is obtained with the introduction of Regions with CNN features (R-CNN) \cite{rcnn}. DNNs, or the most representative CNNs, act in a quite different way from traditional approaches. They have deeper architectures with the capacity to learn more complex features than the shallow ones. Also the expressivity and robust training algorithms allow to learn informative object representations without the need to design features manually \cite{dl_review}.

Since the proposal of R-CNN, a great deal of improved models have been suggested, including Fast R-CNN which jointly optimizes classification and bounding box regression tasks \cite{frcn}, Faster R-CNN which takes an additional sub-network to generate region proposals \cite{Faster} and YOLO which accomplishes object detection via a fixed-grid regression \cite{yolo}. All of them bring different degrees of detection performance improvements over the primary R-CNN and make real-time and accurate object detection become more achievable.

\begin{figure}[!t]
  \centering
  \centerline{\includegraphics[width=0.35\textwidth,height=3.5cm]{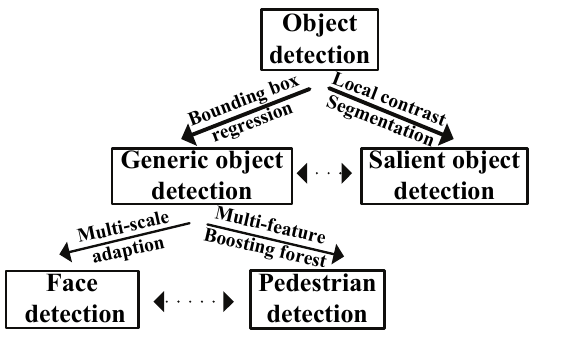}}
\vspace{-0.3cm}
\caption{The application domains of object detection. }
\label{fig:detection_sketch}
\vspace{-0.5cm}
\end{figure}

In this paper, a systematic review is provided to summarise representative models and their different characteristics in several application domains, \textbf{including} generic object detection \cite{rcnn,frcn,Faster}, salient object detection \cite{liu2015predicting,vig2014large}, face detection \cite{faster_face,Joint-cascade,chen2016supervised} and pedestrian detection \cite{ribeiro2016real,sdp}. Their relationships are depicted in Figure \ref{fig:detection_sketch}. Based on basic CNN architectures, generic object detection is achieved with bounding box regression, while salient object detection is accomplished with local contrast enhancement and pixel-level segmentation. Face detection and pedestrian detection are closely related to generic object detection and mainly accomplished with multi-scale adaption and multi-feature fusion/boosting forest, respectively. The dotted lines indicate that the corresponding domains are associated with each other under certain conditions. It should be noticed that the covered domains are diversified. Pedestrian and face images have regular structures, while general objects and scene images have more complex variations in geometric structures and layouts. Therefore, different deep models are required by various images.

There has been a relevant pioneer effort \cite{druzhkov2016survey} which mainly focuses on relevant software tools to implement deep learning techniques for image classification and object detection, but pays little attention on detailing specific algorithms. Different from it, our work not only reviews deep learning based object detection models and algorithms covering different application domains in detail, but also provides their corresponding experimental comparisons and meaningful analyses.

The rest of this paper is organized as follows. In Section 2, a brief introduction on the history of deep learning and the basic architecture of CNN is provided. Generic object detection architectures are presented in Section 3. Then reviews of CNN applied in several specific tasks, including salient object detection, face detection and pedestrian detection, are exhibited in Section 4-6, respectively.  Several promising future directions are proposed in Section 7. At last, some concluding remarks are presented in Section 8.
\section{A Brief Overview of Deep Learning}
Prior to overview on deep learning based object detection approaches, we provide a review on the history of deep learning along with an introduction on the basic architecture and advantages of CNN.
\subsection{The History: Birth, Decline and Prosperity }
Deep models can be referred to as neural networks with deep structures. %\cite{du2006novel,zhao2007mended,du2007shape}
The history of neural networks can date back to 1940s \cite{pitts1947we}, and the original intention was to simulate the human brain system to solve general learning problems in a principled way. It was popular in 1980s and 1990s %\cite{huang1996systematic,huang1999radial}
with the proposal of back-propagation algorithm by Hinton et al. \cite{bp}. However, due to the overfitting of training, lack of large scale training data, limited computation power and insignificance in performance compared with other machine learning tools, neural networks fell out of fashion in early 2000s.

Deep learning has become popular since 2006 \cite{hinton2006reducing}[S7] with a break through in speech recognition \cite{hinton2012deep}. The recovery of deep learning can be attributed to the following factors. % huang2004constructive,huang2005zeroing,zhao2007palmprint,huang2008constructive

\hangafter 0
\hangindent 1em
\noindent
$\bullet$ The emergence of large scale annotated training data, such as ImageNet \cite{imagenet}, to fully exhibit its very large learning capacity;

\hangafter 0
\hangindent 1em
\noindent
$\bullet$ Fast development of high performance parallel computing systems, such as GPU clusters;

\hangafter 0
\hangindent 1em
\noindent
$\bullet$ Significant advances in the design of network structures and training strategies. With unsupervised and layerwise pre-training guided by Auto-Encoder (AE) \cite{deng2010binary} or Restricted Boltzmann Machine (RBM) \cite{rbm}, a good initialization is provided. With dropout and data augmentation, the overfitting problem in training has been relieved \cite{alexnet,dropout}. With batch normalization (BN), the training of very deep neural networks becomes quite efficient \cite{BN}. Meanwhile, various network structures, such as AlexNet \cite{alexnet}, Overfeat \cite{overfeat}, GoogLeNet \cite{goolenet}, VGG \cite{VGG} and ResNet \cite{resnet}, have been extensively studied to improve the performance.

What prompts deep learning to have a huge impact on the entire academic community? It may owe to the contribution of Hinton's group, whose continuous efforts have demonstrated that deep learning would bring a revolutionary breakthrough on grand challenges rather than just obvious improvements on small datasets. Their success results from training a large CNN on 1.2 million labeled images together with a few techniques \cite{alexnet} (e.g., ReLU operation \cite{relu} and `dropout' regularization).
\vspace{-0.3cm}
\subsection{Architecture and Advantages of CNN}

CNN is the most representative model of deep learning \cite{dl_review}. A typical CNN architecture, which is referred to as VGG16, can be found in Fig. S1. Each layer of CNN is known as a feature map. The feature map of the input layer is a 3D matrix of pixel intensities for different color channels (e.g. RGB). The feature map of any internal layer is an induced multi-channel image, whose `pixel' can be viewed as a specific feature. Every neuron is connected with a small portion of adjacent neurons from the previous layer (receptive field). Different types of transformations \cite{alexnet,oquab2014weakly,oquab2014learning} can be conducted on feature maps, such as filtering and pooling. Filtering (convolution) operation convolutes a filter matrix (learned weights) with the values of a receptive field of neurons and takes a nonlinear function (such as sigmoid \cite{sigmoid}, ReLU) to obtain final responses. Pooling operation, such as max pooling, average pooling, L2-pooling and local contrast normalization \cite{LCN}, summaries the responses of a receptive field into one value to produce more robust feature descriptions.

With an interleave between convolution and pooling, an initial feature hierarchy is constructed, which can be fine-tuned in a supervised manner by adding several fully connected (FC) layers to adapt to different visual tasks. According to the tasks involved, the final layer with different activation functions \cite{alexnet} is added to get a specific conditional probability for each output neuron. And the whole network can be optimized on an objective function (e.g. mean squared error or cross-entropy loss) via the stochastic gradient descent (SGD) method. The typical VGG16 has totally 13 convolutional (conv) layers, 3 fully connected layers, 3 max-pooling layers and a softmax classification layer. The conv feature maps are produced by convoluting 3*3 filter windows, and feature map resolutions are reduced with 2 stride max-pooling layers.
%\cite{sgd,huang2012general,jiang2016random}
An arbitrary test image of the same size as training samples can be processed with the trained network. Re-scaling or cropping operations may be needed if different sizes are provided \cite{alexnet}.

The advantages of CNN against traditional methods can be summarised as follows.

\hangafter 0
\hangindent 1em
\noindent
$\bullet$ Hierarchical feature representation, which is the multi-level representations from pixel to high-level semantic features learned by a hierarchical multi-stage structure \cite{rcnn,Kavukcuoglu2010LearningCF}, can be learned from data automatically and hidden factors of input data can be disentangled through multi-level nonlinear mappings.

\hangafter 0
\hangindent 1em
\noindent
$\bullet$ Compared with traditional shallow models, a deeper architecture provides an exponentially increased expressive capability.

\hangafter 0
\hangindent 1em
\noindent
$\bullet$ The architecture of CNN provides an opportunity to jointly optimize several related tasks together (e.g. Fast R-CNN combines classification and bounding box regression into a multi-task leaning manner).

\hangafter 0
\hangindent 1em
\noindent
$\bullet$ Benefitting from the large learning capacity of deep CNNs, some classical computer vision challenges can be recast as high-dimensional data transform problems and solved from a different viewpoint.

Due to these advantages, CNN has been widely applied into many research fields, such as image super-resolution reconstruction \cite{zeiler2010deconvolutional,noh2015learning}, image classification \cite{jia2014caffe,zhao2014plant}, image retrieval \cite{babenko2014neural,wan2014deep}, face recognition \cite{yang2016multi}[S5], pedestrian detection \cite{tome2016deep,subcnn,zhao2017pedestrian} and video analysis \cite{ngiam2011multimodal,wu2015modeling}.
\section{Generic Object Detection}
\begin{figure*}[!t]
  \centering
  \centerline{\includegraphics[width=0.7\textwidth,height=3.8cm]{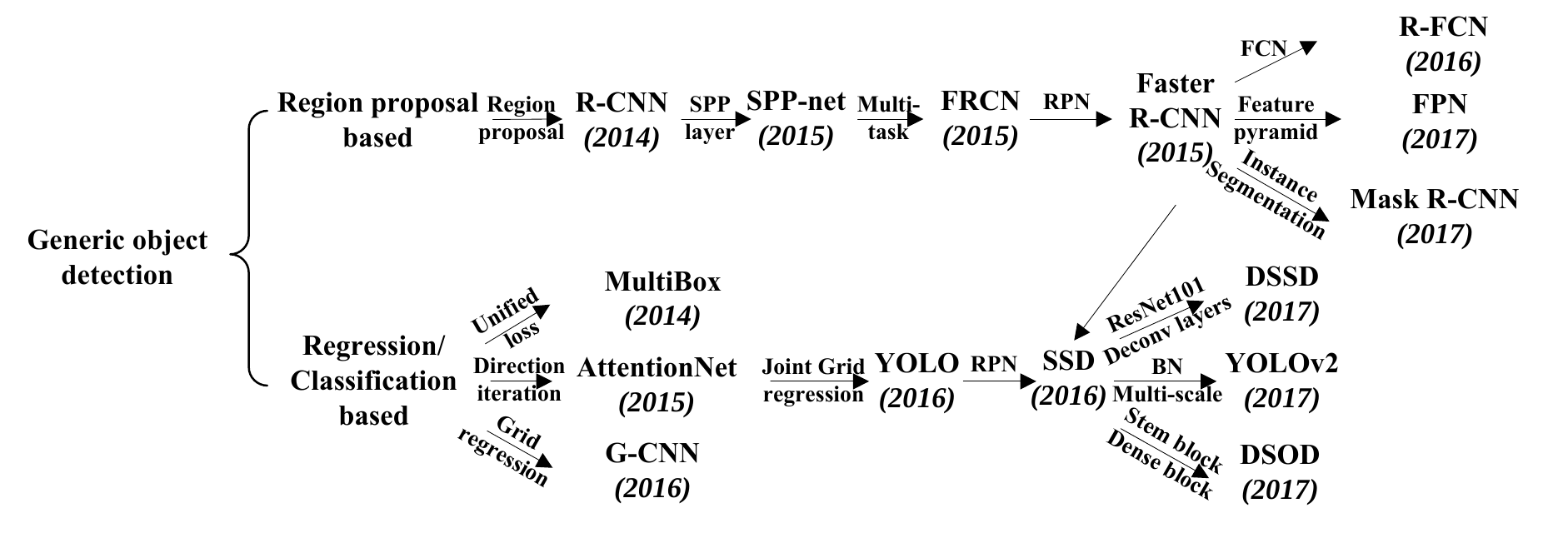}}
\vspace{-0.3cm}
\caption{Two types of frameworks: region proposal based and regression$/$classification based. SPP: Spatial Pyramid Pooling \cite{spp-net}, FRCN: Faster R-CNN \cite{frcn}, RPN: Region Proposal Network \cite{Faster}, FCN: Fully Convolutional Network \cite{rfcn}, BN: Batch Normalization \cite{BN}, Deconv layers: Deconvolution layers \cite{zeiler2010deconvolutional}}.
\label{fig:generic_sketch}
\vspace{-0.5cm}
\end{figure*}

Generic object detection aims at locating and classifying existing objects in any one image, and labeling them with rectangular bounding boxes to show the confidences of existence. The frameworks of generic object detection methods can mainly be categorized into two types (see Figure \ref{fig:generic_sketch}). One follows traditional object detection pipeline, generating region proposals at first and then classifying each proposal into different object categories. The other regards object detection as a regression or classification problem, adopting a unified framework to achieve final results (categories and locations) directly. The region proposal based methods mainly include R-CNN \cite{rcnn}, SPP-net \cite{spp-net}, Fast R-CNN \cite{frcn}, Faster R-CNN \cite{Faster}, R-FCN \cite{rfcn}, FPN \cite{fpn} and Mask R-CNN \cite{mask_rcnn}, some of which are correlated with each other (e.g. SPP-net modifies R-CNN with a SPP layer). The regression$/$classification based methods mainly includes MultiBox \cite{multi-box}, AttentionNet \cite{attentionnet}, G-CNN \cite{gcnn}, YOLO \cite{yolo}, SSD \cite{ssd}, YOLOv2 \cite{yolov2}, DSSD \cite{dssd} and DSOD \cite{DSOD}. The correlations between these two pipelines are bridged by the anchors introduced in Faster R-CNN. Details of these methods are as follows.
\vspace{-0.2cm}

\subsection{Region Proposal Based Framework}
The region proposal based framework, a two-step process, matches the attentional mechanism of human brain to some extent, which gives a coarse scan of the whole scenario firstly and then focuses on regions of interest. Among the pre-related works \cite{overfeat,hinton2011transforming,taylor2011learning}, the most representative one is Overfeat \cite{overfeat}. This model inserts CNN into sliding window method, which predicts bounding boxes directly from locations of the topmost feature map after obtaining the confidences of underlying object categories.
\subsubsection{R-CNN}
\label{sec:rcnn}
\begin{figure}[!t]
  \centering
  \centerline{\includegraphics[width=0.45\textwidth]{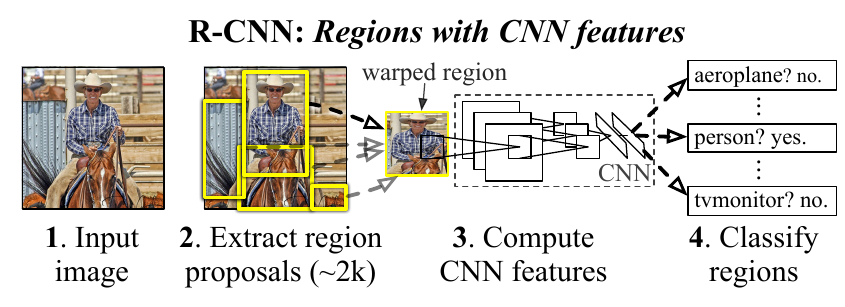}}
\vspace{-0.3cm}
\caption{The flowchart of R-CNN \cite{rcnn}, which consists of 3 stages: (1) extracts bottom-up region proposals, (2) computes features for each proposal using a CNN, and then (3) classifies each region with class-specific linear SVMs. }
\label{fig:r-cnn}
\vspace{-0.5cm}
\end{figure}

It is of significance to improve the quality of candidate bounding boxes and to take a deep architecture to extract high-level features. To solve these problems, R-CNN \cite{rcnn} was proposed by Ross Girshick in 2014 and obtained a mean average precision (mAP) of $53.3\%$ with more than $30\%$ improvement over the previous best result (DPM HSC \cite{DPM_HSC}) on PASCAL VOC 2012. Figure \ref{fig:r-cnn} shows the flowchart of R-CNN, which can be divided into three stages as follows.

\noindent\textbf{Region proposal generation.} The R-CNN adopts selective search \cite{SS} to generate about 2k region proposals for each image. The selective search method relies on simple bottom-up grouping and saliency cues to provide more accurate candidate boxes of arbitrary sizes quickly and to reduce the searching space in object detection \cite{DPM,imagenet}.

\noindent\textbf{CNN based deep feature extraction.} In this stage, each region proposal is warped or cropped into a fixed resolution and the CNN module in \cite{alexnet} is utilized to extract a 4096-dimensional feature as the final representation. Due to large learning capacity, dominant expressive power and hierarchical structure of CNNs, a high-level, semantic and robust feature representation for each region proposal can be obtained.

\noindent\textbf{Classification and localization.} With pre-trained category-specific linear SVMs for multiple classes, different region proposals are scored on a set of positive regions and background (negative) regions. The scored regions are then adjusted with bounding box regression and filtered with a greedy non-maximum suppression (NMS) to produce final bounding boxes for preserved object locations.

When there are scarce or insufficient labeled data, pre-training is usually conducted. Instead of unsupervised pre-training \cite{sermanet2013pedestrian}, R-CNN firstly conducts supervised pre-training on ILSVRC, a very large auxiliary dataset, and then takes a domain-specific fine-tuning. This scheme has been adopted by most of subsequent approaches \cite{frcn,Faster}.

In spite of its improvements over traditional methods and significance in bringing CNN into practical object detection, there are still some disadvantages.

\hangafter 0
\hangindent 1em
\noindent
$\bullet$ Due to the existence of FC layers, the CNN requires a fixed-size (e.g., $227\times227$) input image, which directly leads to the re-computation of the whole CNN for each evaluated region, taking a great deal of time in the testing period.

\hangafter 0
\hangindent 1em
\noindent
$\bullet$ Training of R-CNN is a multi-stage pipeline. At first, a convolutional network (ConvNet) on object proposals is fine-tuned. Then the softmax classifier learned by fine-tuning is replaced by SVMs to fit in with ConvNet features. Finally, bounding-box regressors are trained.

\hangafter 0
\hangindent 1em
\noindent
$\bullet$ Training is expensive in space and time. Features are extracted from different region proposals and stored on the disk. It will take a long time to process a relatively small training set with very deep networks, such as VGG16. At the same time, the storage memory required by these features should also be a matter of concern.

\hangafter 0
\hangindent 1em
\noindent
$\bullet$ Although selective search can generate region proposals with relatively high recalls, the obtained region proposals are still redundant and this procedure is time-consuming (around 2 seconds to extract 2k region proposals).

To solve these problems, many methods have been proposed. GOP \cite{gop} takes a much faster geodesic based segmentation to replace traditional graph cuts. MCG \cite{mcg} searches different scales of the image for multiple hierarchical segmentations and combinatorially groups different regions to produce proposals. Instead of extracting visually distinct segments, the edge boxes method \cite{EB} adopts the idea that objects are more likely to exist in bounding boxes with fewer contours straggling their boundaries. Also some researches tried to re-rank or refine pre-extracted region proposals to remove unnecessary ones and obtained a limited number of valuable ones, such as DeepBox \cite{deepbox} and SharpMask \cite{SharpMask}.

In addition, there are some improvements to solve the problem of inaccurate localization. Zhang et al. \cite{bayes} utilized a bayesian optimization based search algorithm to guide the regressions of different bounding boxes sequentially, and trained class-specific CNN classifiers with a structured loss to penalize the localization inaccuracy explicitly. Saurabh Gupta et al. improved object detection for RGB-D images with semantically rich image and depth features \cite{rgbd}, and learned a new geocentric embedding for depth images to encode each pixel. The combination of object detectors and superpixel classification framework gains a promising result on semantic scene segmentation task. Ouyang et al. proposed a deformable deep CNN (DeepID-Net) \cite{deepid} which introduces a novel deformation constrained pooling (def-pooling) layer to impose geometric penalty on the deformation of various object parts and makes an ensemble of models with different settings.
%By averaging an ensemble of models with different settings and adjusting several key components in the R-CNN pipeline, DeepID-Net achieved a remarkable performance on ILSVRC2014.
Lenc et al. \cite{r-r} provided an analysis on the role of proposal generation in CNN-based detectors and tried to replace this stage with a constant and trivial region generation scheme. The goal is achieved by biasing sampling to match the statistics of the ground truth bounding boxes with K-means clustering. However, more candidate boxes are required to achieve comparable results to those of R-CNN.
%%%%
\subsubsection{SPP-net}
\label{sec:spp}
FC layers must take a fixed-size input. That's why R-CNN chooses to warp or crop each region proposal into the same size. However, the object may exist partly in the cropped region and unwanted geometric distortion may be produced due to the warping operation. These content losses or distortions will reduce recognition accuracy, especially when the scales of objects vary.

To solve this problem, He et al. took the theory of spatial pyramid matching (SPM) \cite{spm,Perronnin2010ImprovingTF} into consideration and proposed a novel CNN architecture named SPP-net \cite{spp-net}. SPM takes several finer to coarser scales to partition the image into a number of divisions and aggregates quantized local features into mid-level representations.

\begin{figure}[!t]
  \centering
  \centerline{\includegraphics[width=.28\textwidth,height=3.8cm]{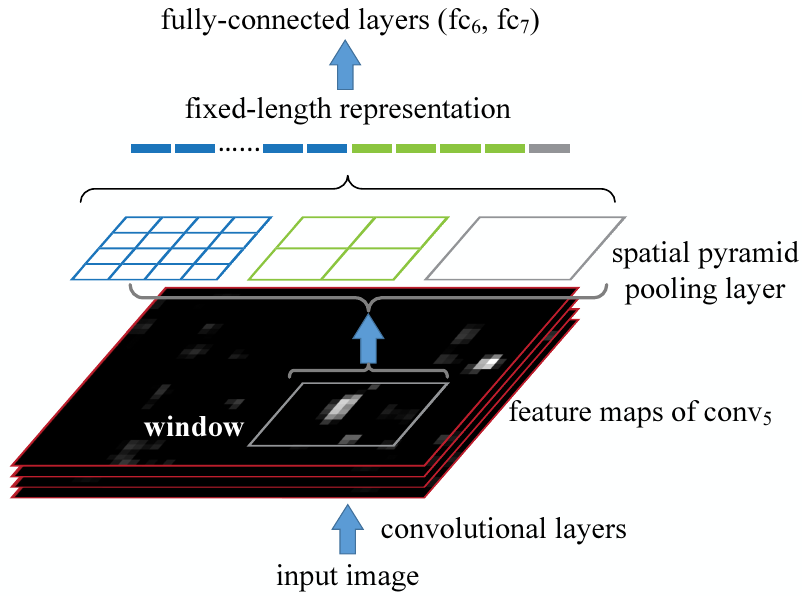}}
\vspace{-0.3cm}
\caption{The architecture of SPP-net for object detection \cite{spp-net}. }
\label{fig:spp_net}
\vspace{-0.5cm}
\end{figure}

The architecture of SPP-net for object detection can be found in Figure \ref{fig:spp_net}. Different from R-CNN, SPP-net reuses feature maps of the 5-th conv layer (conv5) to project region proposals of arbitrary sizes to fixed-length feature vectors. The feasibility of the reusability of these feature maps is due to the fact that the feature maps not only involve the strength of local responses, but also have relationships with their spatial positions \cite{spp-net}. The layer after the final conv layer is referred to as spatial pyramid pooling layer (SPP layer). If the number of feature maps in conv5 is 256, taking a 3-level pyramid, the final feature vector for each region proposal obtained after SPP layer has a dimension of $256\times(1^2+2^2+4^2)=5376$.

SPP-net not only gains better results with correct estimation of different region proposals in their corresponding scales, but also improves detection efficiency in testing period with the sharing of computation cost before SPP layer among different proposals.
\subsubsection{Fast R-CNN}
\label{sec:frcn}
Although SPP-net has achieved impressive improvements in both accuracy and efficiency over R-CNN, it still has some notable drawbacks. SPP-net takes almost the same multi-stage pipeline as R-CNN, including feature extraction, network fine-tuning, SVM training and bounding-box regressor fitting. So an additional expense on storage space is still required. Additionally, the conv layers preceding the SPP layer cannot be updated with the fine-tuning algorithm introduced in \cite{spp-net}. As a result, an accuracy drop of very deep networks is unsurprising. To this end, Girshick \cite{frcn} introduced a multi-task loss on classification and bounding box regression and proposed a novel CNN architecture named Fast R-CNN.

\begin{figure}[!t]
  \centering
  \centerline{\includegraphics[width=.35\textwidth,height=2cm]{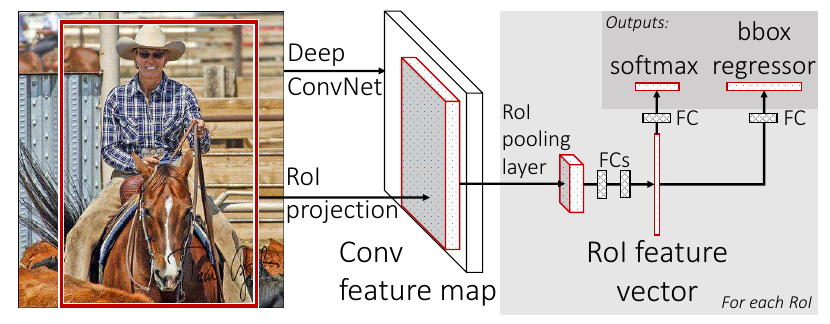}}
\vspace{-0.3cm}
\caption{The architecture of Fast R-CNN \cite{frcn}. }
\label{fig:fast_rcnn}
\vspace{-0.5cm}
\end{figure}

The architecture of Fast R-CNN is exhibited in Figure \ref{fig:fast_rcnn}. Similar to SPP-net, the whole image is processed with conv layers to produce feature maps. Then, a fixed-length feature vector is extracted from each region proposal with a region of interest (RoI) pooling layer. The RoI pooling layer is a special case of the SPP layer, which has only one pyramid level. Each feature vector is then fed into a sequence of FC layers before finally branching into two sibling output layers. One output layer is responsible for producing softmax probabilities for all $C+1$ categories ($C$ object classes plus one `background' class) and the other output layer encodes refined bounding-box positions with four real-valued numbers. All parameters in these procedures (except the generation of region proposals) are optimized via a multi-task loss in an end-to-end way.

The multi-tasks loss $L$ is defined as below to jointly train classification and bounding-box regression,
\begin{equation}
\label{eqn:loss_fast_rcnn}
L(p,u,t^{u},v)=L_{cls}(p,u)+\lambda[u\geq1]L_{loc}(t^{u},v)
\end{equation}
where $L_{cls}(p,u)=-\log p_{u}$ calculates the log loss for ground truth class $u$ and $p_{u}$ is driven from the discrete probability distribution $p=(p_{0},\cdots,p_{C})$ over the $C+1$ outputs from the last FC layer. $L_{loc}(t^{u},v)$ is defined over the predicted offsets $t^{u} = (t_{x}^{u},t_{y}^{u},t_{w}^{u},t_{h}^{u})$ and ground-truth bounding-box regression targets $v = (v_{x},v_{y},v_{w},v_{h})$, where $x,y,w,h$ denote the two coordinates of the box center, width, and height, respectively. Each $t^{u}$ adopts the parameter settings in \cite{rcnn} to specify an object proposal with a log-space height/width shift and scale-invariant translation. The Iverson bracket indicator function $[u\geq1]$ is employed to omit all background RoIs. To provide more robustness against outliers and eliminate the sensitivity in exploding gradients, a smooth $L_{1}$ loss is adopted to fit bounding-box regressors as below
\begin{equation}
\label{eqn:loc_loss_fast_rcnn}
L_{loc}(t^{u},v)=\sum_{i\in{x,y,w,h}}smooth_{L_{1}}(t_{i}^{u}-v_{i})
\end{equation}
where
\begin{equation}
smooth_{L_{1}}(x)=\begin{cases}
0.5x^{2} & if\left|x\right|<1\\
\left|x\right|-0.5 & otherwise
\end{cases}
\end{equation}

To accelerate the pipeline of Fast R-CNN, another two tricks are of necessity. On one hand, if training samples (i.e. RoIs) come from different images, back-propagation through the SPP layer becomes highly inefficient. Fast R-CNN samples mini-batches hierarchically, namely $N$ images sampled randomly at first and then $R/N$ RoIs sampled in each image, where $R$ represents the number of RoIs. Critically, computation and memory are shared by RoIs from the same image in the forward and backward pass. On the other hand, much time is spent in computing the FC layers during the forward pass \cite{frcn}. The truncated Singular Value Decomposition (SVD) \cite{svd} can be utilized to compress large FC layers and to accelerate the testing procedure.

In the Fast R-CNN, regardless of region proposal generation, the training of all network layers can be processed in a single-stage with a multi-task loss. It saves the additional expense on storage space, and improves both accuracy and efficiency with more reasonable training schemes.
\subsubsection{Faster R-CNN}
\label{sec:faster}
Despite the attempt to generate candidate boxes with biased sampling \cite{r-r}, state-of-the-art object detection networks mainly rely on additional methods, such as selective search and Edgebox, to generate a candidate pool of isolated region proposals. Region proposal computation is also a bottleneck in improving efficiency. To solve this problem, Ren et al. introduced an additional Region Proposal Network (RPN) \cite{Faster_tpami,Faster}, which acts in a nearly cost-free way by sharing full-image conv features with detection network.

RPN is achieved with a fully-convolutional network, which has the ability to predict object bounds and scores at each position simultaneously. Similar to \cite{SS}, RPN takes an image of arbitrary size to generate a set of rectangular object proposals. RPN operates on a specific conv layer with the preceding layers shared with object detection network.

\begin{figure}[!t]
  \centering
  \centerline{\includegraphics[width=.4\textwidth,height=3.0cm]{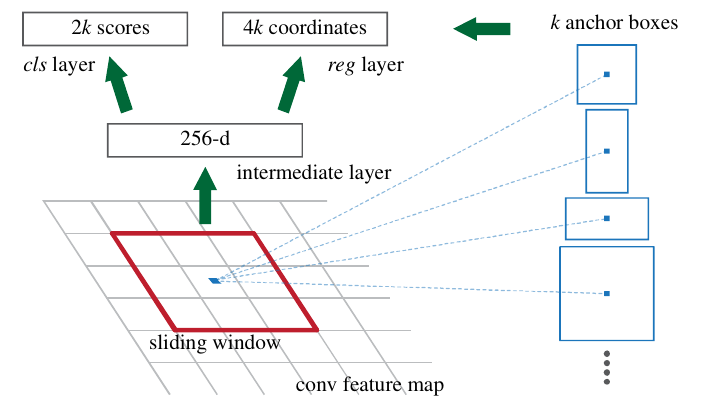}}
\vspace{-0.3cm}
\caption{The RPN in Faster R-CNN \cite{Faster}. K predefined anchor boxes are convoluted with each sliding window to produce fixed-length vectors which are taken by cls and reg layer to obtain corresponding outputs. }
\label{fig:rpn}
\vspace{-0.5cm}
\end{figure}

The architecture of RPN is shown in Figure \ref{fig:rpn}. The network slides over the conv feature map and fully connects to an $n\times n$ spatial window. A low dimensional vector (512-d for VGG16) is obtained in each sliding window and fed into two sibling FC layers, namely box-classification layer (cls) and box-regression layer (reg). This architecture is implemented with an $n\times n$ conv layer followed by two sibling $1\times 1$ conv layers. To increase non-linearity, ReLU is applied to the output of the $n\times n$ conv layer.

The regressions towards true bounding boxes are achieved by comparing proposals relative to reference boxes (anchors). In the Faster R-CNN, anchors of 3 scales and 3 aspect ratios are adopted. The loss function is similar to (\ref{eqn:loss_fast_rcnn}).
\begin{equation}
L(p_{i},t_{i})=\frac{1}{N_{cls}}\sum_{i}L_{cls}(p_{i},p_{i}^{*})+\lambda \frac{1}{N_{reg}}\sum_{i}p_{i}^{*}L_{reg}(t_{i},t_{i}^{*})
\end{equation}
where $p_{i}$ shows the predicted probability of the $i$-th anchor being an object. The ground truth label $p_{i}^{*}$ is 1 if the anchor is positive, otherwise 0. $t_{i}$ stores 4 parameterized coordinates of the predicted bounding box while $t_{i}^{*}$ is related to the ground-truth box overlapping with a positive anchor. $L_{cls}$ is a binary log loss and $L_{reg}$ is a smoothed $L_{1}$ loss similar to (\ref{eqn:loc_loss_fast_rcnn}). These two terms are normalized with the mini-batch size ($N_{cls}$) and the number of anchor locations ($N_{reg}$), respectively. In the form of fully-convolutional networks, Faster R-CNN can be trained end-to-end by back-propagation and SGD in an alternate training manner.

With the proposal of Faster R-CNN, region proposal based CNN architectures for object detection can really be trained in an end-to-end way. Also a frame rate of 5 FPS (Frame Per Second) on a GPU is achieved with state-of-the-art object detection accuracy on PASCAL VOC 2007 and 2012. However, the alternate training algorithm is very time-consuming and RPN produces object-like regions (including backgrounds) instead of object instances and is not skilled in dealing with objects with extreme scales or shapes.
\subsubsection{R-FCN}
\label{sec:rfcn}
Divided by the RoI pooling layer, a prevalent family \cite{frcn,Faster} of deep networks for object detection are composed of two subnetworks: a shared fully convolutional subnetwork (independent of RoIs) and an unshared RoI-wise subnetwork. This decomposition originates from pioneering classification architectures (e.g. AlexNet \cite{alexnet} and VGG16 \cite{VGG}) which consist of a convolutional subnetwork and several FC layers separated by a specific spatial pooling layer.

Recent state-of-the-art image classification networks, such as Residual Nets (ResNets) \cite{resnet} and GoogLeNets \cite{goolenet,szegedy2016rethinking}, are fully convolutional. To adapt to these architectures, it's natural to construct a fully convolutional object detection network without RoI-wise subnetwork. However, it turns out to be inferior with such a naive solution \cite{resnet}. This inconsistence is due to the dilemma of respecting translation variance in object detection compared with increasing translation invariance in image classification. In other words, shifting an object inside an image should be indiscriminative in image classification while any translation of an object in a bounding box may be meaningful in object detection. A manual insertion of the RoI pooling layer into convolutions can break down translation invariance at the expense of additional unshared region-wise layers. So Li et al. \cite{rfcn} proposed a region-based fully convolutional networks (R-FCN, Fig. S2).

Different from Faster R-CNN, for each category, the last conv layer of R-FCN produces a total of $k^{2}$ position-sensitive score maps with a fixed grid of $k\times k$ firstly and a position-sensitive RoI pooling layer is then appended to aggregate the responses from these score maps. Finally, in each RoI, $k^2$ position-sensitive scores are averaged to produce a $C+1$-d vector and softmax responses across categories are computed. Another $4k^2$-d conv layer is appended to obtain class-agnostic bounding boxes.

With R-FCN, more powerful classification networks can be adopted to accomplish object detection in a fully-convolutional architecture by sharing nearly all the layers, and state-of-the-art results are obtained on both PASCAL VOC and Microsoft COCO \cite{ms-coco} datasets at a test speed of 170ms per image.
\subsubsection{FPN}
\label{sec:fpn}
\begin{figure}[!t]
  \centering
  \centerline{\includegraphics[width=.46\textwidth,height=4.3cm]{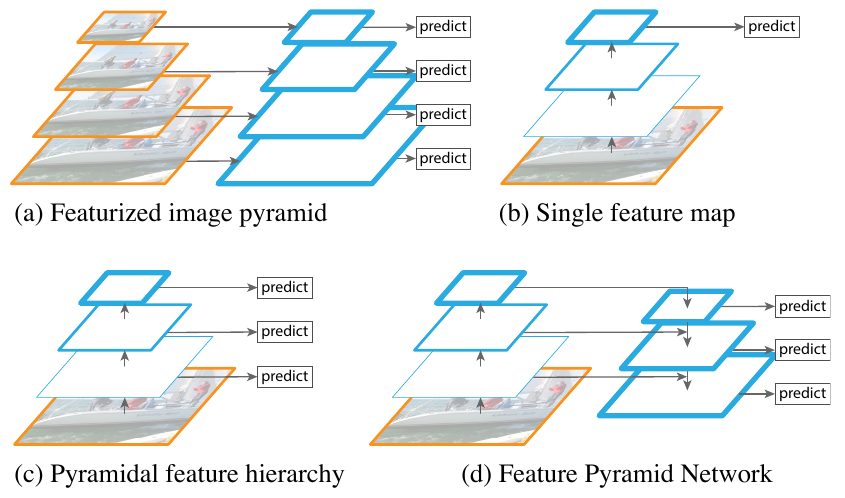}}
\vspace{-0.3cm}
\caption{The main concern of FPN \cite{fpn}. (a) It is slow to use an image pyramid to build a feature pyramid. (b) Only single scale features is adopted for faster detection. (c) An alternative to the featurized image pyramid is to reuse the pyramidal feature hierarchy computed by a ConvNet. (d) FPN integrates both (b) and (c). Blue outlines indicate feature maps and thicker outlines denote semantically stronger features.}
\label{fig:fpn}
\vspace{-0.5cm}
\end{figure}
Feature pyramids built upon image pyramids (featurized image pyramids) have been widely applied in many object detection systems to improve scale invariance \cite{DPM,spp-net} (Figure \ref{fig:fpn}(a)). However, training time and memory consumption increase rapidly. To this end, some techniques take only a single input scale to represent high-level semantics and increase the robustness to scale changes (Figure \ref{fig:fpn}(b)), and image pyramids are built at test time which results in an inconsistency between train/test-time inferences \cite{frcn,Faster}. The in-network feature hierarchy in a deep ConvNet produces feature maps of different spatial resolutions while introduces large semantic gaps caused by different depths (Figure \ref{fig:fpn}(c)). To avoid using low-level features, pioneer works \cite{ssd,ION} usually build the pyramid starting from middle layers or just sum transformed feature responses, missing the higher-resolution maps of the feature hierarchy.

Different from these approaches, FPN \cite{fpn} holds an architecture with a bottom-up pathway, a top-down pathway and several lateral connections to combine low-resolution and semantically strong features with high-resolution and semantically weak features (Figure \ref{fig:fpn}(d)). The bottom-up pathway, which is the basic forward backbone ConvNet, produces a feature hierarchy by downsampling the corresponding feature maps with a stride of 2. The layers owning the same size of output maps are grouped into the same network stage and the output of the last layer of each stage is chosen as the reference set of feature maps to build the following top-down pathway.

To build the top-down pathway, feature maps from higher network stages are upsampled at first and then enhanced with those of the same spatial size from the bottom-up pathway via lateral connections. A $1 \times 1$ conv layer is appended to the upsampled map to reduce channel dimensions and the mergence is achieved by element-wise addition. Finally, a $3 \times 3$ convolution is also appended to each merged map to reduce the aliasing effect of upsampling and the final feature map is generated. This process is iterated until the finest resolution map is generated.

As feature pyramid can extract rich semantics from all levels and be trained end-to-end with all scales, state-of-the-art representation can be obtained without sacrificing speed and memory. Meanwhile, FPN is independent of the backbone CNN architectures and can be applied to different stages of object detection (e.g. region proposal generation) and to many other computer vision tasks (e.g. instance segmentation).
\subsubsection{Mask R-CNN}
\label{sec:mask}
\begin{figure}[!t]
  \centering
  \centerline{\includegraphics[width=.4\textwidth,height=2cm]{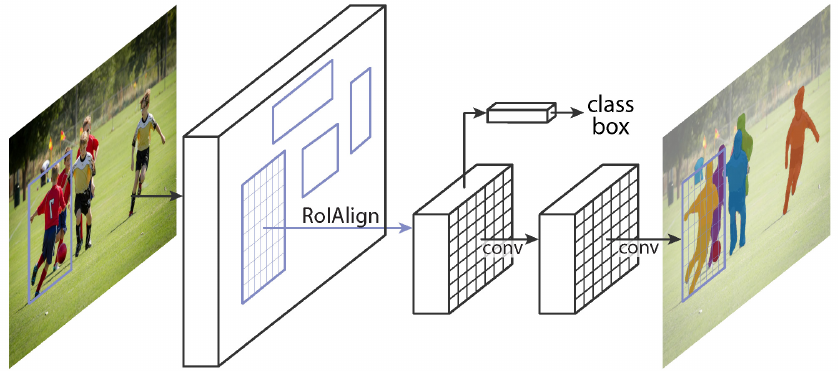}}
\vspace{-0.3cm}
\caption{The Mask R-CNN framework for instance segmentation \cite{mask_rcnn}.}
\label{fig:mask}
\vspace{-0.5cm}
\end{figure}

Instance segmentation \cite{Arnab2017Pixelwise} is a challenging task which requires detecting all objects in an image and segmenting each instance (semantic segmentation \cite{dai2016instance}). These two tasks are usually regarded as two independent processes. And the multi-task scheme will create spurious edge and exhibit systematic errors on overlapping instances \cite{Li2017FullyCI}. To solve this problem, parallel to the existing branches in Faster R-CNN for classification and bounding box regression, the Mask R-CNN \cite{mask_rcnn} adds a branch to predict segmentation masks in a pixel-to-pixel manner (Figure \ref{fig:mask}).

Different from the other two branches which are inevitably collapsed into short output vectors by FC layers, the segmentation mask branch encodes an $m \times m$ mask to maintain the explicit object spatial layout. This kind of fully convolutional representation requires fewer parameters but is more accurate than that of \cite{dai2016instance}. Formally, besides the two losses in (\ref{eqn:loss_fast_rcnn}) for classification and bounding box regression, an additional loss for segmentation mask branch is defined to reach a multi-task loss. An this loss is only associated with ground-truth class and relies on the classification branch to predict the category.

Because RoI pooling, the core operation in Faster R-CNN, performs a coarse spatial quantization for feature extraction, misalignment is introduced between the RoI and the features. It affects classification little because of its robustness to small translations. However, it has a large negative effect on pixel-to-pixel mask prediction. To solve this problem, Mask R-CNN adopts a simple and quantization-free layer, namely RoIAlign, to preserve the explicit per-pixel spatial correspondence faithfully. RoIAlign is achieved by replacing the harsh quantization of RoI pooling with bilinear interpolation \cite{Jaderberg2015SpatialTN}, computing the exact values of the input features at four regularly sampled locations in each RoI bin. In spite of its simplicity, this seemingly minor change improves mask accuracy greatly, especially under strict localization metrics.

Given the Faster R-CNN framework, the mask branch only adds a small computational burden and its cooperation with other tasks provides complementary information for object detection. As a result, Mask R-CNN is simple to implement with promising instance segmentation and object detection results. In a word, Mask R-CNN is a flexible and efficient framework for instance-level recognition, which can be easily generalized to other tasks (e.g. human pose estimation \cite{Cao2017RealtimeM2}[S4]) with minimal modification.
\subsubsection{Multi-task Learning, Multi-scale Representation and Contextual Modelling}
\label{sec:m_m_c}
Although the Faster R-CNN gets promising results with several hundred proposals, it still struggles in small-size object detection and localization, mainly due to the coarseness of its feature maps and limited information provided in particular candidate boxes. The phenomenon is more obvious on the Microsoft COCO dataset which consists of objects at a broad range of scales, less prototypical images, and requires more precise localization. To tackle these problems, it is of necessity to accomplish object detection with multi-task learning \cite{stuffnet}, multi-scale representation \cite{ION} and context modelling \cite{hypernet} to combine complementary information from multiple sources.

\textbf{Multi-task Learning} learns a useful representation for multiple correlated tasks from the same input \cite{pentina2015curriculum,yim2015rotating}. Brahmbhatt et al. introduced conv features trained for object segmentation and `stuff' (amorphous categories such as ground and water) to guide accurate object detection of small objects (StuffNet) \cite{stuffnet}. Dai et al. \cite{dai2016instance} presented Multitask Network Cascades of three networks, namely class-agnostic region proposal generation, pixel-level instance segmentation and regional instance classification. Li et al. incorporated the weakly-supervised object segmentation cues and region-based object detection into a multi-stage architecture to fully exploit the learned segmentation features \cite{li2016multi}.

\textbf{Multi-scale Representation} combines activations from multiple layers with skip-layer connections to provide semantic information of different spatial resolutions \cite{fpn}. Cai et al. proposed the MS-CNN \cite{mscnn} to ease the inconsistency between the sizes of objects and receptive fields with multiple scale-independent output layers. Yang et al. investigated two strategies, namely scale-dependent pooling (SDP) and layerwise cascaded rejection classifiers (CRC), to exploit appropriate scale-dependent conv features \cite{sdp}. Kong et al. proposed the HyperNet to calculate the shared features between RPN and object detection network by aggregating and compressing hierarchical feature maps from different resolutions into a uniform space \cite{hypernet}.

\textbf{Contextual Modelling} improves detection performance by exploiting features from or around RoIs of different support regions and resolutions to deal with occlusions and local similarities \cite{ION}. Zhu et al. proposed the SegDeepM to exploit object segmentation which reduces the dependency on initial candidate boxes with Markov Random Field \cite{segdeepm}. Moysset et al. took advantage of 4 directional 2D-LSTMs \cite{lstm_scene} to convey global context between different local regions and reduced trainable parameters with local parameter-sharing \cite{moysset2016learning}. Zeng et al. proposed a novel GBD-Net by introducing gated functions to control message transmission between different support regions \cite{GBD-net}.
%To release the heavy dependence on massive annotated data and processing power

\textbf{The Combination} incorporates different components above into the same model to improve detection performance further. Gidaris et al. proposed the Multi-Region CNN (MR-CNN) model \cite{mr-cnn} to capture different aspects of an object, the distinct appearances of various object parts and semantic segmentation-aware features.
%However, the improvements on localization performance are mainly due to increasing context at multiple scales around objects rather than focusing on different object parts and the network cannot be trained end-to-end.
To obtain contextual and multi-scale representations, Bell et al. proposed the Inside-Outside Net (ION) by exploiting information both inside and outside the RoI \cite{ION} with spatial recurrent neural networks \cite{rnn} and skip pooling \cite{hypernet}.
%However, in this architecture, the hidden transitions of recurrent layers are set to identity and hard to be learned.
Zagoruyko et al. proposed the MultiPath architecture by introducing three modifications to the Fast R-CNN  \cite{multipath}, including multi-scale skip connections \cite{ION}, a modified foveal structure \cite{mr-cnn} and a novel loss function summing different IoU losses.
\subsubsection{Thinking in Deep Learning based Object Detection}
Apart from the above approaches, there are still many important factors for continued progress.

There is a large imbalance between the number of annotated objects and background examples. To address this problem, Shrivastava et al. proposed an effective online mining algorithm (OHEM) \cite{OHEM} for automatic selection of the hard examples, which leads to a more effective and efficient training.

Instead of concentrating on feature extraction, Ren et al. made a detailed analysis on object classifiers \cite{NOC}, and found that it is of particular importance for object detection to construct a deep and convolutional per-region classifier carefully, especially for ResNets \cite{resnet} and GoogLeNets \cite{goolenet}.

%Subcategory-aware Convolutional Neural Networks for Object Proposals and Detection (KITTI, VOC 07)
Traditional CNN framework for object detection is not skilled in handling significant scale variation, occlusion or truncation, especially when only 2D object detection is involved. To address this problem, Xiang et al. proposed a novel subcategory-aware region proposal network \cite{subcnn}, which guides the generation of region proposals with subcategory information related to object poses and jointly optimize object detection and subcategory classification.
%As a result, the state-of-the-art performance on both object detection and pose estimation is obtained.

%Factors in Finetuning Deep Model for Object Detection with Long-tail Distribution£º
Ouyang et al. found that the samples from different classes follow a longtailed distribution \cite{longtail}, which indicates that different classes with distinct numbers of samples have different degrees of impacts on feature learning. To this end, objects are firstly clustered into visually similar class groups, and then a hierarchical feature learning scheme is adopted to learn deep representations for each group separately.

In order to minimize computational cost and achieve the state-of-the-art performance, with the `deep and thin' design principle and following the pipeline of Fast R-CNN, Hong et al. proposed the architecture of PVANET \cite{pvanet}, which adopts some building blocks including concatenated ReLU \cite{crelu}, Inception \cite{goolenet}, and HyperNet \cite{hypernet} to reduce the expense on multi-scale feature extraction and trains the network with batch normalization \cite{BN}, residual connections \cite{resnet}, and learning rate scheduling based on plateau detection \cite{resnet}. The PVANET achieves the state-of-the-art performance and can be processed in real time on Titan X GPU (21 FPS).
\vspace{-0.3cm}
\subsection{Regression$/$Classification Based Framework}
\label{sec:poineer}
Region proposal based frameworks are composed of several correlated stages, including region proposal generation, feature extraction with CNN, classification and bounding box regression, which are usually trained separately. Even in recent end-to-end module Faster R-CNN, an alternative training is still required to obtain shared convolution parameters between RPN and detection network. As a result, the time spent in handling different components becomes the bottleneck in real-time application.

One-step frameworks based on global regression/classification, mapping straightly from image pixels to bounding box coordinates and class probabilities, can reduce time expense. We firstly reviews some pioneer CNN models, and then focus on two significant frameworks, namely You only look once (YOLO) \cite{yolo} and Single Shot MultiBox Detector (SSD) \cite{ssd}.
\subsubsection{Pioneer Works}
%Deep Neural Networks for Object Detection£º
Previous to YOLO and SSD, many researchers have already tried to model object detection as a regression or classification task.

Szegedy et al. formulated object detection task as a DNN-based regression \cite{binary_mask}, generating a binary mask for the test image and extracting detections with a simple bounding box inference. However, the model has difficulty in handling overlapping objects, and bounding boxes generated by direct upsampling is far from perfect.

Pinheiro et al. proposed a CNN model with two branches: one generates class agnostic segmentation masks and the other predicts the likelihood of a given patch centered on an object \cite{pinheiro2015learning}. Inference is efficient since class scores and segmentation can be obtained in a single model with most of the CNN operations shared.

%multi-box:
%the pioneer multi-layer approaches based on bottom-up classless segmentation
Erhan et al. proposed regression based MultiBox to produce scored class-agnostic region proposals \cite{multi-box,MSC-MultiBox}. A unified loss was introduced to bias both localization and confidences of multiple components to predict the coordinates of class-agnostic bounding boxes. However, a large quantity of additional parameters are introduced to the final layer.

%Attention-net(art):
Yoo et al. adopted an iterative classification approach to handle object detection and proposed an impressive end-to-end CNN architecture named AttentionNet \cite{attentionnet}. Starting from the top-left (TL) and bottom-right (BR) corner of an image, AttentionNet points to a target object by generating quantized weak directions and converges to an accurate object boundary box with an ensemble of iterative predictions. However, the model becomes quite inefficient when handling multiple categories with a progressive two-step procedure.

%Gcnn£¨objective function£¬drawbacks for multi-box and attention-net£¬rcnn minus r£©£º
Najibi et al. proposed a proposal-free iterative grid based object detector (G-CNN), which models object detection as finding a path from a fixed grid to boxes tightly surrounding the objects \cite{gcnn}. Starting with a fixed multi-scale bounding box grid, G-CNN trains a regressor to move and scale elements of the grid towards objects iteratively. However, G-CNN has a difficulty in dealing with small or highly overlapping objects.
\subsubsection{YOLO}
\label{sec:yolo}
\begin{figure}[!t]
  \centering
  \centerline{\includegraphics[width=.38\textwidth,height=3.5cm]{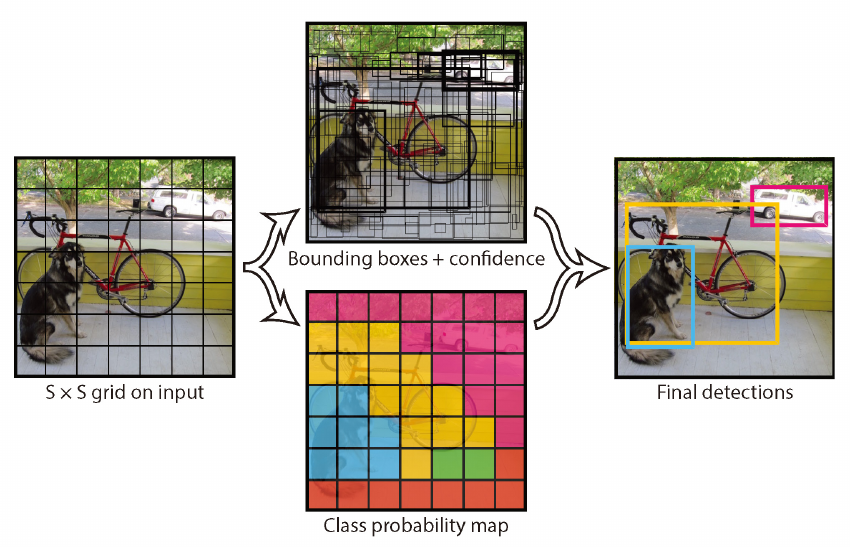}}
\vspace{-0.3cm}
\caption{Main idea of YOLO \cite{yolo}. }
\label{fig:yolo}
\vspace{-0.5cm}
\end{figure}

Redmon et al.\cite{yolo} proposed a novel framework called YOLO, which makes use of the whole topmost feature map to predict both confidences for multiple categories and bounding boxes. The basic idea of YOLO is exhibited in Figure \ref{fig:yolo}. YOLO divides the input image into an S $\times$ S grid and each grid cell is responsible for predicting the object centered in that grid cell. Each grid cell predicts $B$ bounding boxes and their corresponding confidence scores. Formally, confidence scores are defined as $Pr(Object)*IOU^{truth}_{pred}$, which indicates how likely there exist objects ($Pr(Object)\geq0$) and shows confidences of its prediction ($IOU^{truth}_{pred}$). At the same time, regardless of the number of boxes, $C$ conditional class probabilities ($Pr(Class_{i}|Object)$) should also be predicted in each grid cell. It should be noticed that only the contribution from the grid cell containing an object is calculated.

At test time, class-specific confidence scores for each box are achieved by multiplying the individual box confidence predictions with the conditional class probabilities as follows.
\begin{equation}
\begin{gathered}
Pr(Object)*IOU_{pred}^{truth}*Pr(Class_{i}|Object)\\
=Pr(Class_{i})*IOU_{pred}^{truth}
\end{gathered}
\end{equation}
where the existing probability of class-specific objects in the box and the fitness between the predicted box and the object are both taken into consideration.

\begin{figure*}[!t]
  \centering
  \centerline{\includegraphics[width=.7\textwidth,height=3.3cm]{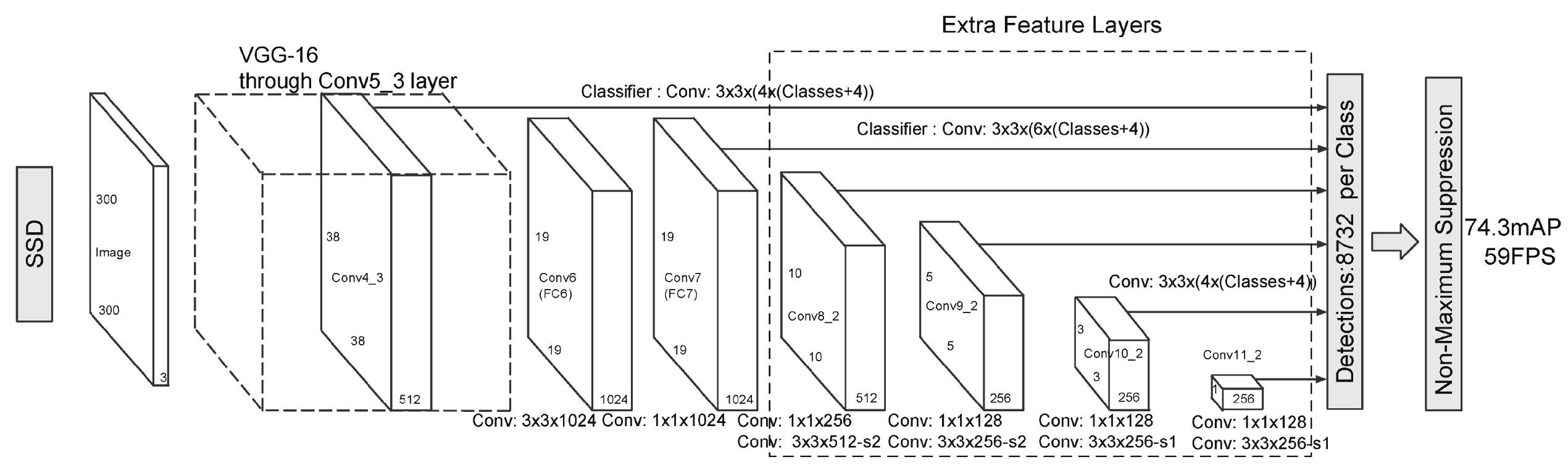}}
\vspace{-0.3cm}
\caption{The architecture of SSD 300 \cite{ssd}. SSD adds several feature layers to the end of VGG16 backbone network to predict the offsets to default anchor boxes and their associated confidences. Final detection results are obtained by conducting NMS on multi-scale refined bounding boxes. }
\label{fig:ssd_art}
\vspace{-0.5cm}
\end{figure*}

During training, the following loss function is optimized,
\begin{equation}
\begin{aligned}
\lambda_{coord}\sum_{i=0}^{S^2}\sum_{j=0}^{B}\mathbbm{1}_{ij}^{obj}&\left[(x_{i}-\hat{x_{i}})^2+(y_{i}-\hat{y_{i}})^2\right]\\
+\lambda_{coord}\sum_{i=0}^{S^2}\sum_{j=0}^{B} &\mathbbm{1}_{ij}^{obj}\left[\left(\sqrt{w_{i}}-\sqrt{\hat{w_{i}}})^2+(\sqrt{h_{i}}-\sqrt{\hat{h_{i}}}\right)^2\right]\\
&+\sum_{i=0}^{S^2}\sum_{j=0}^{B}\mathbbm{1}_{ij}^{obj}\left(C_{i}-\hat{C_{i}}\right)^2\\
+\lambda_{noobj}&\sum_{i=0}^{S^2}\sum_{j=0}^{B}\mathbbm{1}_{ij}^{noobj}\left(C_{i}-\hat{C_{i}}\right)^2\\
&+\sum_{i=0}^{S^2}\mathbbm{1}_{i}^{obj}\sum_{c\in classes}(p_{i}(c)-\hat{p}_{i}(c))^2
\end{aligned}
\end{equation}
In a certain cell $i$, $(x_{i},y_{i})$ denote the center of the box relative to the bounds of the grid cell, $(w_{i},h_{i})$ are the normalized width and height relative to the image size, $C_{i}$ represents confidence scores, $\mathbbm{1}_{i}^{obj}$ indicates the existence of objects and $\mathbbm{1}_{ij}^{obj}$ denotes that the prediction is conducted by the $j$th bounding box predictor. Note that only when an object is present in that grid cell, the loss function penalizes classification errors. Similarly, when the predictor is `responsible' for the ground truth box (i.e. the highest IoU of any predictor in that grid cell is achieved), bounding box coordinate errors are penalized.

The YOLO consists of 24 conv layers and 2 FC layers, of which some conv layers construct ensembles of inception modules with 1 $\times$ 1 reduction layers followed by 3 $\times$ 3 conv layers.
%Taking PASCAL VOC as an example, an experimental setting of $S = 7, B = 2$ and $C = 20$ is applied, which results in an output of 7 $\times$7$\times$30 tensor predicting different grid cells.
The network can process images in real-time at 45 FPS and a simplified version Fast YOLO can reach 155 FPS with better results than other real-time detectors. Furthermore, YOLO produces fewer false positives on background, which makes the cooperation with Fast R-CNN become possible. An improved version, YOLOv2, was later proposed in \cite{yolov2}, which adopts several impressive strategies, such as BN, anchor boxes, dimension cluster and multi-scale training.
\subsubsection{SSD}
\label{sec:ssd}
YOLO has a difficulty in dealing with small objects in groups, which is caused by strong spatial constraints imposed on bounding box predictions \cite{yolo}. Meanwhile, YOLO struggles to generalize to objects in new/unusual aspect ratios/ configurations and produces relatively coarse features due to multiple downsampling operations.

Aiming at these problems, Liu et al. proposed a Single Shot MultiBox Detector (SSD) \cite{ssd}, which was inspired by the anchors adopted in MultiBox \cite{multi-box}, RPN \cite{Faster} and multi-scale representation \cite{ION}. Given a specific feature map, instead of fixed grids adopted in YOLO, the SSD takes advantage of a set of default anchor boxes with different aspect ratios and scales to discretize the output space of bounding boxes. To handle objects with various sizes, the network fuses predictions from multiple feature maps with different resolutions .

The architecture of SSD is demonstrated in Figure \ref{fig:ssd_art}. Given the VGG16 backbone architecture, SSD adds several feature layers to the end of the network, which are responsible for predicting the offsets to default boxes with different scales and aspect ratios and their associated confidences. The network is trained with a weighted sum of localization loss (e.g. Smooth L1) and confidence loss (e.g. Softmax), which is similar to (\ref{eqn:loss_fast_rcnn}). Final detection results are obtained by conducting NMS on multi-scale refined bounding boxes.

Integrating with hard negative mining, data augmentation and a larger number of carefully chosen default anchors, SSD significantly outperforms the Faster R-CNN in terms of accuracy on PASCAL VOC and COCO, while being three times faster. The SSD300 (input image size is $300\times300$) runs at 59 FPS, which is more accurate and efficient than YOLO. However, SSD is not skilled at dealing with small objects, which can be relieved by adopting better feature extractor backbone (e.g. ResNet101), adding deconvolution layers with skip connections to introduce additional large-scale context \cite{dssd} and designing better network structure (e.g. Stem Block and Dense Block) \cite{DSOD}.
\vspace{-0.3cm}
\subsection{Experimental Evaluation}
\begin{table*}[!tb]
\renewcommand\arraystretch{0.9}
\caption{An overview of prominent generic object detection architectures.  }
\vspace{-0.2cm}
\setlength{\tabcolsep}{4pt}
\label{tab:overview}
\centering
\tiny
\begin{threeparttable}[b]
\begin{tabular}{c|c|c|c|c|c|c|c|c|c}
\toprule[2pt]
Framework & Proposal & Multi-scale Input & Learning Method &Loss Function &Softmax Layer &End-to-end Train  &Platform &Language \\
\midrule[2pt]
R-CNN\cite{rcnn} 	&Selective Search &- &SGD,BP	&Hinge loss (classification),Bounding box regression &+ &- &Caffe &Matlab\tabularnewline
%\hline
SPP-net\cite{spp-net} & EdgeBoxes	& +	& SGD & Hinge loss (classification),Bounding box regression & + & - & Caffe & Matlab\tabularnewline
Fast RCNN\cite{frcn} & Selective Search & + & SGD	& Class Log loss+bounding box regression & + & - & Caffe & Python\tabularnewline
Faster R-CNN\cite{Faster}& RPN & + & SGD & Class Log loss+bounding box regression & + & + & Caffe & Python/Matlab\tabularnewline
R-FCN\cite{rfcn} & RPN	& +	& SGD & Class Log loss+bounding box regression & - & + & Caffe & Matlab\tabularnewline
\multirow{2}{*}{Mask R-CNN\cite{mask_rcnn} }& \multirow{2}{*}{RPN} & \multirow{2}{*}{+} & \multirow{2}{*}{SGD} & Class Log loss+bounding box regression & \multirow{2}{*}{+} & \multirow{2}{*}{+}  & \multirow{2}{*}{TensorFlow/Keras} &\multirow{2}{*}{Python} \tabularnewline
 &  &  &  & +Semantic sigmoid loss &  & &  & \tabularnewline
FPN\cite{fpn} & RPN & + & Synchronized SGD & Class Log loss+bounding box regression & + & + & TensorFlow & Python\tabularnewline
\multirow{2}{*}{YOLO\cite{yolo}}& \multirow{2}{*}{-} & \multirow{2}{*}{-} & \multirow{2}{*}{SGD} & Class sum-squared error loss+bounding box regression & \multirow{2}{*}{+} & \multirow{2}{*}{+}  & \multirow{2}{*}{Darknet} &\multirow{2}{*}{C} \tabularnewline
 &  &  &  & +object confidence+background confidence	& & &  & \tabularnewline
SSD\cite{ssd} & -	& -	& SGD & Class softmax loss+bounding box regression & - & +  & Caffe & C++\tabularnewline
\multirow{2}{*}{YOLOv2\cite{yolov2}}& \multirow{2}{*}{-} & \multirow{2}{*}{-} & \multirow{2}{*}{SGD} & Class sum-squared error loss+bounding box regression & \multirow{2}{*}{+} & \multirow{2}{*}{+}  & \multirow{2}{*}{Darknet} &\multirow{2}{*}{C} \tabularnewline
 &  &  &  & +object confidence+background confidence &  & &  & \tabularnewline
\bottomrule[2pt]
\end{tabular}
\begin{tablenotes}
  \item[*] `+' denotes that corresponding techniques are employed while `-' denotes that this technique is not considered. It should be noticed that R-CNN and SPP-net can not be trained end-to-end with a multi-task loss while the other architectures are based on multi-task joint training. As most of these architectures are re-implemented on different platforms with various programming languages, we only list the information associated with the versions by the referenced authors.
\end{tablenotes}
\end{threeparttable}
\end{table*}

We compare various object detection methods on three benchmark datasets, including PASCAL VOC 2007 \cite{voc2007}, PASCAL VOC 2012 \cite{voc2012} and Microsoft COCO \cite{ms-coco}. The evaluated approaches include R-CNN \cite{rcnn}, SPP-net \cite{spp-net}, Fast R-CNN \cite{frcn}, NOC \cite{NOC}, Bayes \cite{bayes}, MR-CNN$\&$S-CNN \cite{mscnn}, Faster R-CNN \cite{Faster}, HyperNet \cite{hypernet}, ION \cite{ION}, MS-GR \cite{li2016multi}, StuffNet \cite{stuffnet}, SSD300 \cite{ssd}, SSD512 \cite{ssd}, OHEM \cite{OHEM}, SDP+CRC \cite{sdp}, GCNN \cite{gcnn}, SubCNN \cite{subcnn}, GBD-Net \cite{GBD-net}, PVANET \cite{pvanet}, YOLO \cite{yolo}, YOLOv2 \cite{yolov2}, R-FCN \cite{rfcn}, FPN \cite{fpn}, Mask R-CNN \cite{mask_rcnn}, DSSD \cite{dssd} and DSOD \cite{DSOD}. If no specific instructions for the adopted framework are provided, the utilized model is a VGG16 \cite{VGG} pretrained on 1000-way ImageNet classification task \cite{imagenet}. Due to the limitation of paper length, we only provide an overview, including proposal, learning method, loss function, programming language and platform, of the prominent architectures in Table \ref{tab:overview}. Detailed experimental settings, which can be found in the original papers, are missed. In addition to the comparisons of detection accuracy, another comparison is provided to evaluate their test consumption on PASCAL VOC 2007.

\begin{table*}[!t]
\renewcommand\arraystretch{0.8}
\vspace{-0.5cm}
\caption{Comparative results on VOC 2007 test set (\%). }
\vspace{-0.2cm}
\setlength{\tabcolsep}{4pt}
\label{tab:voc07}
\centering
\tiny
\begin{threeparttable}[b]
\begin{tabular}{c|c|cccccccccccccccccccc|c}
\toprule[2pt]
Methods & Trained on &areo &bike &bird &boat &bottle &bus &car &cat &chair &cow &table &dog &horse &mbike &person &plant &sheep &sofa &train &tv &mAP \tabularnewline
\midrule[2pt]
R-CNN (Alex)\cite{rcnn} &07 &68.1	&72.8	&56.8	&43.0	&36.8	&66.3	&74.2	&67.6	&34.4	&63.5	&54.5	&61.2	&69.1	&68.6	&58.7	&33.4	&62.9	&51.1	&62.5	&68.6	&58.5 \tabularnewline
R-CNN(VGG16)\cite{rcnn} 	&07	&73.4	&77.0	&63.4	&45.4	&44.6	&75.1	&78.1	&79.8	&40.5	&73.7	&62.2	&79.4	&78.1	&73.1	&64.2	&35.6	&66.8	&67.2	&70.4	&71.1	&66.0 \tabularnewline
SPP-net(ZF)\cite{spp-net} &07	&68.5	&71.7	&58.7	&41.9	&42.5	&67.7	&72.1	&73.8	&34.7	&67.0	&63.4	&66.0	&72.5	&71.3	&58.9	&32.8	&60.9	&56.1	&67.9	&68.8	&60.9  \tabularnewline
GCNN\cite{gcnn}	&07	&68.3	&77.3	&68.5	&52.4	&38.6	&78.5	&79.5	&81.0	&47.1	&73.6	&64.5	&77.2	&80.5	&75.8	&66.6	&34.3	&65.2	&64.4	&75.6	&66.4	&66.8 \tabularnewline
Bayes\cite{bayes}	&07	&74.1	&83.2	&67.0	&50.8	&51.6	&76.2	&81.4	&77.2	&48.1	&78.9	&65.6	&77.3	&78.4	&75.1	&70.1	&41.4	&69.6	&60.8	&70.2	&73.7	&68.5 \tabularnewline
Fast R-CNN\cite{frcn}	&07+12	&77.0	&78.1	&69.3	&59.4	&38.3	&81.6	&78.6	&86.7	&42.8	&78.8	&68.9	&84.7	&82.0	&76.6	&69.9	&31.8	&70.1	&74.8	&80.4	&70.4	&70.0 \tabularnewline
SDP+CRC\cite{sdp} &07	&76.1	&79.4	&68.2	&52.6	&46.0	&78.4	&78.4	&81.0	&46.7	&73.5	&65.3	&78.6	&81.0	&76.7	&77.3	&39.0	&65.1	&67.2	&77.5	&70.3	&68.9 \tabularnewline
SubCNN\cite{subcnn} &07	&70.2	&80.5	&69.5	&60.3	&47.9	&79.0	&78.7	&84.2	&48.5	&73.9	&63.0	&82.7	&80.6	&76.0	&70.2	&38.2	&62.4	&67.7	&77.7	&60.5	&68.5  \tabularnewline
StuffNet30\cite{stuffnet} &07	&72.6	&81.7	&70.6	&60.5	&53.0	&81.5	&83.7	&83.9	&52.2	&78.9	&70.7	&85.0	&85.7	&77.0	&78.7	&42.2	&73.6	&69.2	&79.2	&73.8	&72.7  \tabularnewline
NOC\cite{NOC} &07+12	&76.3	&81.4	&74.4	&61.7	&60.8	&84.7	&78.2	&82.9	&53.0	&79.2	&69.2	&83.2	&83.2	&78.5	&68.0	&45.0	&71.6	&76.7	&82.2	&75.7	&73.3  \tabularnewline
MR-CNN$\&$S-CNN\cite{mr-cnn} &07+12	&80.3	&84.1	&78.5	&70.8	&\bf 68.5	&88.0	&85.9	&87.8	&60.3	&85.2	&73.7	&87.2	&86.5	&85.0	&76.4	&48.5	&76.3	&75.5	&85.0	&81.0	&78.2\tabularnewline
HyperNet\cite{hypernet} &07+12	&77.4	&83.3	&75.0	&69.1	&62.4	&83.1	&87.4	&87.4	&57.1	&79.8	&71.4	&85.1	&85.1	&80.0	&79.1	&51.2	&79.1	&75.7	&80.9	&76.5	&76.3  \tabularnewline
MS-GR\cite{li2016multi}	&07+12	&80.0	&81.0	&77.4	&72.1	&64.3	&88.2	&88.1	&88.4	&64.4	&85.4	&73.1	&87.3	&87.4	&85.1	&79.6	&50.1	&78.4	&79.5	&86.9	&75.5	&78.6 \tabularnewline
OHEM+Fast R-CNN\cite{OHEM}	&07+12	&80.6	&85.7	&79.8	&69.9	&60.8	&88.3	&87.9	&89.6	&59.7	&85.1	&76.5	&87.1	&87.3	&82.4	&78.8	&53.7	&80.5	&78.7	&84.5	&80.7	&78.9 \tabularnewline
ION\cite{ION} &07+12+S	&80.2	&85.2	&78.8	&70.9	&62.6	&86.6	&86.9	&\bf 89.8	&61.7	&86.9	&76.5	&\bf 88.4	&87.5	&83.4	&80.5	&52.4	&78.1	&77.2	&86.9	&\bf 83.5 &79.2\tabularnewline
Faster R-CNN\cite{Faster}	&07	&70.0	&80.6	&70.1	&57.3	&49.9	&78.2	&80.4	&82.0	&52.2	&75.3	&67.2	&80.3	&79.8	&75.0	&76.3	&39.1	&68.3	&67.3	&81.1	&67.6	&69.9  \tabularnewline
Faster R-CNN\cite{Faster}	&07+12	&76.5	&79.0	&70.9	&65.5	&52.1	&83.1	&84.7	&86.4	&52.0	&81.9	&65.7	&84.8	&84.6	&77.5	&76.7	&38.8	&73.6	&73.9	&83.0	&72.6	&73.2 \tabularnewline
Faster R-CNN\cite{Faster}	&07+12+COCO	&84.3	&82.0	&77.7	&68.9	&65.7	&88.1	&88.4	&88.9	&63.6	&86.3	&70.8	&85.9	&87.6	&80.1	&82.3	&53.6	&80.4	&75.8	&86.6	&78.9	&78.8 \tabularnewline
SSD300\cite{ssd} &07+12+COCO	&80.9	&86.3	&79.0	&\bf 76.2	&57.6	&87.3	&88.2	&88.6	&60.5	&85.4	&\bf 76.7	&87.5	&\bf 89.2	&84.5	&81.4	&55.0	&81.9	&\bf 81.5	&85.9	&78.9	&79.6  \tabularnewline
SSD512\cite{ssd} &07+12+COCO	&\bf 86.6	&\bf 88.3	&\bf 82.4	&76.0	&66.3	&\bf 88.6	&\bf 88.9	&89.1	&\bf 65.1	&\bf 88.4	&73.6	&86.5	&88.9	&\bf 85.3	&\bf 84.6	&\bf 59.1	&\bf 85.0	&80.4	&\bf 87.4	&81.2	&\bf 81.6\tabularnewline
\bottomrule[2pt]
\end{tabular}
\begin{tablenotes}
  \item[*] `07': VOC2007 trainval, `07+12': union of VOC2007 and VOC2012 trainval, `07+12+COCO': trained on COCO trainval35k at first and then fine-tuned on 07+12. The S in ION `07+12+S' denotes SBD segmentation labels.
\end{tablenotes}
\end{threeparttable}
\end{table*}
\begin{table*}[!t]
\renewcommand\arraystretch{0.8}
\vspace{-0.5cm}
\caption{Comparative results on VOC 2012 test set (\%). }
\vspace{-0.2cm}
\setlength{\tabcolsep}{4pt}
\label{tab:voc12}
\centering
\tiny
\begin{threeparttable}[b]
\begin{tabular}{c|c|cccccccccccccccccccc|c}
\toprule[2pt]
Methods & Trained on &areo &bike &bird &boat &bottle &bus &car &cat &chair &cow &table &dog &horse &mbike &person &plant &sheep &sofa &train &tv &mAP \tabularnewline
\midrule[2pt]
R-CNN(Alex)\cite{rcnn} &12	&71.8	&65.8	&52.0	&34.1	&32.6	&59.6	&60.0	&69.8	&27.6	&52.0	&41.7	&69.6	&61.3	&68.3	&57.8	&29.6	&57.8	&40.9	&59.3	&54.1	&53.3 \tabularnewline
R-CNN(VGG16)\cite{rcnn}	&12	&79.6	&72.7	&61.9	&41.2	&41.9	&65.9	&66.4	&84.6	&38.5	&67.2	&46.7	&82.0	&74.8	&76.0	&65.2	&35.6	&65.4	&54.2	&67.4	&60.3	&62.4  \tabularnewline
Bayes\cite{bayes}	&12	&82.9	&76.1	&64.1	&44.6	&49.4	&70.3	&71.2	&84.6	&42.7	&68.6	&55.8	&82.7	&77.1	&79.9	&68.7	&41.4	&69.0	&60.0	&72.0	&66.2	&66.4  \tabularnewline
Fast R-CNN\cite{frcn} &07++12	&82.3	&78.4	&70.8	&52.3	&38.7	&77.8	&71.6	&89.3	&44.2	&73.0	&55.0	&87.5	&80.5	&80.8	&72.0	&35.1	&68.3	&65.7	&80.4	&64.2	&68.4 \tabularnewline
SutffNet30\cite{stuffnet}	&12	&83.0	&76.9	&71.2	&51.6	&50.1	&76.4	&75.7	&87.8	&48.3	&74.8	&55.7	&85.7	&81.2	&80.3	&79.5	&44.2	&71.8	&61.0	&78.5	&65.4	&70.0 \tabularnewline
NOC\cite{NOC}	&07+12	&82.8	&79.0	&71.6	&52.3	&53.7	&74.1	&69.0	&84.9	&46.9	&74.3	&53.1	&85.0	&81.3	&79.5	&72.2	&38.9	&72.4	&59.5	&76.7	&68.1	&68.8 \tabularnewline
MR-CNN$\&$S-CNN\cite{mr-cnn}	&07++12	&85.5	&82.9	&76.6	&57.8	&62.7	&79.4	&77.2	&86.6	&55.0	&79.1	&62.2	&87.0	&83.4	&84.7	&78.9	&45.3	&73.4	&65.8	&80.3	&74.0	&73.9 \tabularnewline
HyperNet\cite{hypernet}	&07++12	&84.2	&78.5	&73.6	&55.6	&53.7	&78.7	&79.8	&87.7	&49.6	&74.9	&52.1	&86.0	&81.7	&83.3	&81.8	&48.6	&73.5	&59.4	&79.9	&65.7	&71.4 \tabularnewline
OHEM+Fast R-CNN\cite{OHEM}	&07++12+coco	&90.1	&87.4	&79.9	&65.8	&66.3	&86.1	&85.0	&92.9	&62.4	&83.4	&69.5	&90.6	&88.9	&88.9	&83.6	&59.0	&82.0	&74.7	&88.2	&77.3	&80.1  \tabularnewline
ION\cite{ION}	&07+12+S	&87.5	&84.7	&76.8	&63.8	&58.3	&82.6	&79.0	&90.9	&57.8	&82.0	&64.7	&88.9	&86.5	&84.7	&82.3	&51.4	&78.2	&69.2	&85.2	&73.5	&76.4 \tabularnewline
Faster R-CNN\cite{Faster}	&07++12	&84.9	&79.8	&74.3	&53.9	&49.8	&77.5	&75.9	&88.5	&45.6	&77.1	&55.3	&86.9	&81.7	&80.9	&79.6	&40.1	&72.6	&60.9	&81.2	&61.5	&70.4 \tabularnewline
Faster R-CNN\cite{Faster}	&07++12+coco	&87.4	&83.6	&76.8	&62.9	&59.6	&81.9	&82.0	&91.3	&54.9	&82.6	&59.0	&89.0	&85.5	&84.7	&84.1	&52.2	&78.9	&65.5	&85.4	&70.2	&75.9  \tabularnewline
YOLO\cite{yolo} &07++12  &77.0 &67.2 &57.7 &38.3 &22.7 &68.3 &55.9 &81.4 &36.2 &60.8 &48.5 &77.2 &72.3 &71.3 &63.5 &28.9 &52.2 &54.8 &73.9 &50.8  &57.9\tabularnewline
 YOLO+Fast R-CNN\cite{yolo} &07++12  &83.4 &78.5 &73.5 &55.8 &43.4 &79.1 &73.1 &89.4 &49.4 &75.5 &57.0 &87.5 &80.9 &81.0 &74.7 &41.8 &71.5 &68.5 &82.1 &67.2  &70.7\tabularnewline
YOLOv2\cite{yolov2}	&07++12+coco	&88.8	&87.0	&77.8	&64.9	&51.8	&85.2	&79.3	&93.1	&64.4	&81.4	&\bf 70.2	&91.3	&88.1	&87.2	&81.0	&57.7	&78.1	&71.0	&88.5	&76.8	&78.2 \tabularnewline
SSD300\cite{ssd}	&07++12+coco	&91.0	&86.0	&78.1	&65.0	&55.4	&84.9	&84.0	&93.4	&62.1	&83.6	&67.3	&91.3	&88.9	&88.6	&85.6	&54.7	&83.8	&77.3	&88.3	&76.5	&79.3 \tabularnewline
SSD512\cite{ssd}	&07++12+coco	&91.4	&88.6	&82.6	&71.4	&63.1	& \bf 87.4	&88.1	&93.9	&66.9	&86.6	&66.3	&92.0	&91.7	&90.8	&88.5	&60.9	&87.0	&75.4	&90.2	&80.4	&82.2 \tabularnewline
R-FCN (ResNet101)\cite{frcn}	&07++12+coco	&\bf 92.3	&\bf 89.9	& \bf 86.7	&\bf 74.7	&\bf 75.2	&86.7	&\bf 89.0	&\bf 95.8	&\bf 70.2	&\bf 90.4	&66.5	&\bf 95.0	&\bf 93.2	&\bf 92.1	&\bf 91.1	&\bf 71.0	&\bf 89.7	&\bf 76.0	&\bf 92.0	&\bf 83.4	&\bf 85.0
\tabularnewline
\bottomrule[2pt]
\end{tabular}
\begin{tablenotes}
  \item[*] `07++12': union of VOC2007 trainval and test and VOC2012 trainval. `07++12+COCO': trained on COCO trainval35k at first then fine-tuned on 07++12.
\end{tablenotes}
\end{threeparttable}
\vspace{-0.6cm}
\end{table*}
\subsubsection{PASCAL VOC 2007/2012}
PASCAL VOC 2007 and 2012 datasets consist of 20 categories. The evaluation terms are Average Precision (AP) in each single category and mean Average Precision (mAP) across all the 20 categories. Comparative results are exhibited in Table \ref{tab:voc07} and \ref{tab:voc12}, from which the following remarks can be obtained.

\hangafter 0
\hangindent 1em
\noindent
$\bullet$ If incorporated with a proper way, more powerful backbone CNN models can definitely improve object detection performance (the comparison among R-CNN with AlexNet, R-CNN with VGG16 and SPP-net with ZF-Net \cite{zfnet}).

\hangafter 0
\hangindent 1em
\noindent
$\bullet$ With the introduction of SPP layer (SPP-net), end-to-end multi-task architecture (FRCN) and RPN (Faster R-CNN), object detection performance is improved gradually and apparently.

\hangafter 0
\hangindent 1em
\noindent
$\bullet$ Due to large quantities of trainable parameters, in order to obtain multi-level robust features, data augmentation is very important for deep learning based models (Faster R-CNN with `07' ,`07+12' and `07+12+coco').

\hangafter 0
\hangindent 1em
\noindent
$\bullet$ Apart from basic models, there are still many other factors affecting object detection performance, such as multi-scale and multi-region feature extraction (e.g. MR-CNN), modified classification networks (e.g. NOC), additional information from other correlated tasks (e.g. StuffNet, HyperNet), multi-scale representation (e.g. ION) and mining of hard negative samples (e.g. OHEM).

\hangafter 0
\hangindent 1em
\noindent
$\bullet$ As YOLO is not skilled in producing object localizations of high IoU, it obtains a very poor result on VOC 2012. However, with the complementary information from Fast R-CNN (YOLO+FRCN) and the aid of other strategies, such as anchor boxes, BN and fine grained features, the localization errors are corrected (YOLOv2).

\hangafter 0
\hangindent 1em
\noindent
$\bullet$ By combining many recent tricks and modelling the whole network as a fully convolutional one, R-FCN achieves a more obvious improvement of detection performance over other approaches.
\subsubsection{Microsoft COCO}
Microsoft COCO is composed of 300,000 fully segmented images, in which each image has an average of 7 object instances from a total of 80 categories. As there are a lot of less iconic objects with a broad range of scales and a stricter requirement on object localization, this dataset is more challenging than PASCAL 2012. Object detection performance is evaluated by AP computed under different degrees of IoUs and on different object sizes. The results are shown in Table \ref{tab:coco}.
\begin{table}[!tb]
\renewcommand\arraystretch{0.8}
\caption{Comparative results on Microsoft COCO test dev set (\%).  }
\vspace{-0.2cm}
\setlength{\tabcolsep}{1.2pt}
\label{tab:coco}
\centering
\tiny
\begin{threeparttable}[b]
\begin{tabular}{c|c|cccccccccccc}
\toprule[2pt]
Methods & Trained on & 0.5:0.95 & 0.5 &0.75 &S &M &L &1 &10 &100&S &M &L \tabularnewline
\midrule[2pt]
Fast R-CNN\cite{frcn}	&train	&20.5	&39.9	&19.4	&4.1	&20.0	&35.8	&21.3	&29.4	&30.1	&7.3	&32.1	&52.0 \tabularnewline
ION\cite{ION}	&train	&23.6	&43.2	&23.6	&6.4	&24.1	&38.3	&23.2	&32.7	&33.5	&10.1	&37.7	&53.6 \tabularnewline
NOC+FRCN(VGG16)\cite{NOC}	&train	&21.2	&41.5	&19.7	&-	&-	&-	&-	&-	&-	&-	&-	&- \tabularnewline
NOC+FRCN(Google)\cite{NOC}	&train	&24.8	&44.4	&25.2	&-	&-	&-	&-	&-	&-	&-	&-	&- \tabularnewline
NOC+FRCN (ResNet101)\cite{NOC}	&train	&27.2	&48.4	&27.6	&-	&-	&-	&-	&-	&-	&-	&-	&- \tabularnewline
GBD-Net\cite{GBD-net}	&train	&27.0	&45.8	&-	&-	&-	&-	&-	&-	&-	&-	&-	&- \tabularnewline
OHEM+FRCN\cite{OHEM}	&train	&22.6	&42.5	&22.2	&5.0	&23.7	&34.6	&-	&-	&-	&-	&-	&-\tabularnewline
OHEM+FRCN*\cite{OHEM}	&train	&24.4	&44.4	&24.8	&7.1	&26.4	&37.9	&-	&-	&-	&-	&-	&-\tabularnewline
OHEM+FRCN*\cite{OHEM}	&trainval	&25.5	&45.9	&26.1	&7.4	&27.7	&38.5	&-	&-	&-	&-	&-	&- \tabularnewline
Faster R-CNN\cite{Faster}	&trainval	&24.2	&45.3	&23.5	&7.7	&26.4	&37.1	&23.8	&34.0	&34.6	&12.0	&38.5	&54.4 \tabularnewline
YOLOv2\cite{yolov2}	&trainval35k	&21.6	&44.0	&19.2	&5.0	&22.4	&35.5	&20.7	&31.6	&33.3	&9.8	&36.5	&54.4 \tabularnewline
SSD300\cite{ssd}	&trainval35k	&23.2	&41.2	&23.4	&5.3	&23.2	&39.6	&22.5	&33.2	&35.3	&9.6	&37.6	&56.5 \tabularnewline
SSD512\cite{ssd}	&trainval35k	&26.8	&46.5	&27.8	&9.0	&28.9	&41.9	&24.8	&37.5	&39.8	&14.0	&43.5	&59.0 \tabularnewline
R-FCN (ResNet101)\cite{rfcn}	&trainval	&29.2	&51.5	&-	&10.8	&32.8	&45.0	&-	&-	&-	&-&	-	&- \tabularnewline
R-FCN*(ResNet101)\cite{rfcn}	&trainval	&29.9	&51.9	&-	&10.4	&32.4	&43.3	&-	&-	&-	&-	&-	&- \tabularnewline
R-FCN**(ResNet101)\cite{rfcn}	&trainval	&31.5	&53.2	&-	& 14.3	&35.5	&44.2	&-	&-	&-	&-	&-	&- \tabularnewline
Multi-path\cite{multipath}	&trainval	& 33.2	&51.9	& 36.3	& 13.6 & 37.2	& 47.8	& 29.9	& 46.0	& 48.3	& 23.4	& 56.0	& 66.4 \tabularnewline
FPN (ResNet101)\cite{fpn}	&trainval35k	&36.2	&59.1	& 39.0	&18.2	&39.0	&48.2	&-	&-	&-	&-&	-	&- \tabularnewline
Mask (ResNet101+FPN)\cite{mask_rcnn}	&trainval35k	&38.2	&60.3	&41.7	&20.1	&41.1	&50.2	&-	&-	&-	&-&	-	&- \tabularnewline
Mask (ResNeXt101+FPN)\cite{mask_rcnn}	&trainval35k	&\bf 39.8	& \bf 62.3	&\bf 43.4	& \bf 22.1  &\bf  43.2 &\bf 51.2	&-	&-	&-	&-	&-	&- \tabularnewline
 DSSD513 (ResNet101)\cite{dssd}	&trainval35k	&33.2 	& 53.3	&35.2	& 13.0 &35.4 &51.1	&28.9	&43.5	&46.2	&21.8	&49.1	&66.4 \tabularnewline
 DSOD300\cite{DSOD} &trainval	&29.3	& 47.3	&30.6	& 9.4  & 31.5 &47.0	&27.3	&40.7	&43.0	&16.7	&47.1	&65.0 \tabularnewline
\bottomrule[2pt]
\end{tabular}
\begin{tablenotes}
  \item[*] FRCN*: Fast R-CNN with multi-scale training, R-FCN*: R-FCN with multi-scale training, R-FCN**: R-FCN with multi-scale training and testing, Mask: Mask R-CNN.
\end{tablenotes}
\end{threeparttable}
\vspace{-0.7cm}
\end{table}

Besides similar remarks to those of PASCAL VOC, some other conclusions can be drawn as follows from Table \ref{tab:coco}.

\hangafter 0
\hangindent 1em
\noindent
$\bullet$ Multi-scale training and test are beneficial in improving object detection performance, which provide additional information in different resolutions (R-FCN). FPN and DSSD provide some better ways to build feature pyramids to achieve multi-scale representation. The complementary information from other related tasks is also helpful for accurate object localization (Mask R-CNN with instance segmentation task).

\hangafter 0
\hangindent 1em
\noindent
$\bullet$ Overall, region proposal based methods, such as Faster R-CNN and R-FCN, perform better than regression/classfication based approaches, namely YOLO and SSD, due to the fact that quite a lot of localization errors are produced by regression/classfication based approaches.

\hangafter 0
\hangindent 1em
\noindent
$\bullet$ Context modelling is helpful to locate small objects, which provides additional information by consulting nearby objects and surroundings (GBD-Net and multi-path).

\hangafter 0
\hangindent 1em
\noindent
$\bullet$ Due to the existence of a large number of nonstandard small objects, the results on this dataset are much worse than those of VOC 2007/2012. With the introduction of other powerful frameworks (e.g. ResNeXt \cite{resnext}) and useful strategies (e.g. multi-task learning \cite{mask_rcnn,deformable}), the performance can be improved.

\hangafter 0
\hangindent 1em
\noindent
$\bullet$ The success of DSOD in training from scratch stresses the importance of network design to release the requirements for perfect pre-trained classifiers on relevant tasks and large numbers of annotated samples.
\subsubsection{Timing Analysis}
\begin{table}[!tb]
\renewcommand\arraystretch{0.9}
\caption{Comparison of testing consumption on VOC 07 test set. }
\vspace{-0.2cm}
\label{tab:time_voc07}
\centering
\tiny
\begin{threeparttable}[b]
\begin{tabular}{c|c|c|c|c}
\toprule[2pt]
Methods & Trained on & mAP(\%) & Test time(sec/img) & Rate(FPS) \tabularnewline
\midrule[2pt]
SS+R-CNN\cite{rcnn}	&07	&66.0	&32.84	&0.03  \tabularnewline
SS+SPP-net\cite{spp-net}	&07	&63.1	&2.3	&0.44  \tabularnewline
SS+FRCN\cite{frcn}	&07+12	&66.9	&1.72	&0.6  \tabularnewline
SDP+CRC\cite{sdp}	&07	&68.9	&0.47	&2.1 \tabularnewline
SS+HyperNet*\cite{hypernet}	&07+12	&76.3	&0.20	&5 \tabularnewline
MR-CNN$\&$S-CNN\cite{mr-cnn}	&07+12	&78.2	&30	&0.03  \tabularnewline
ION\cite{ION}	&07+12+S	&79.2	&1.92	&0.5  \tabularnewline
Faster R-CNN(VGG16)\cite{Faster}	&07+12	&73.2	&0.11	&9.1  \tabularnewline
Faster R-CNN(ResNet101)\cite{Faster}	&07+12	& \bf 83.8	&2.24	&0.4  \tabularnewline
YOLO\cite{yolo}	&07+12	&63.4	& \bf 0.02	&\bf 45  \tabularnewline
SSD300\cite{ssd}	&07+12	&74.3	& \bf 0.02	&\bf 46  \tabularnewline
SSD512\cite{ssd}	&07+12	&76.8	&0.05	&19 \tabularnewline
R-FCN(ResNet101)\cite{rfcn}	&07+12+coco	&83.6	&0.17	&5.9 \tabularnewline
YOLOv2(544*544)\cite{yolov2}	&07+12	&78.6	&0.03	&40  \tabularnewline
DSSD321(ResNet101)\cite{dssd}	&07+12	&78.6	&0.07	&13.6 \tabularnewline
DSOD300\cite{DSOD}	&07+12+coco	&81.7	&0.06	&17.4 \tabularnewline
PVANET+\cite{pvanet}	&07+12+coco	&\bf 83.8	&0.05	&21.7 \tabularnewline
PVANET+(compress)\cite{pvanet}	&07+12+coco	&82.9	&0.03	& 31.3 \tabularnewline
\bottomrule[2pt]
\end{tabular}
\begin{tablenotes}
  \item[*] SS: Selective Search \cite{rcnn}, SS*: `fast mode' Selective Search \cite{frcn}, HyperNet*: the speed up version of HyperNet and PAVNET+ (compresss): PAVNET with additional bounding box voting and compressed fully convolutional layers.
\end{tablenotes}
\end{threeparttable}
\vspace{-0.5cm}
\end{table}
Timing analysis (Table \ref{tab:time_voc07}) is conducted on Intel i7-6700K CPU with a single core and NVIDIA Titan X GPU. Except for `SS' which is processed with CPU, the other procedures related to CNN are all evaluated on GPU. From Table \ref{tab:time_voc07}, we can draw some conclusions as follows.

\hangafter 0
\hangindent 1em
\noindent
$\bullet$ By computing CNN features on shared feature maps (SPP-net), test consumption is reduced largely. Test time is further reduced with the unified multi-task learning (FRCN) and removal of additional region proposal generation stage (Faster R-CNN). It's also helpful to compress the parameters of FC layers with SVD \cite{svd} (PAVNET and FRCN).

\hangafter 0
\hangindent 1em
\noindent
$\bullet$ It takes additional test time to extract multi-scale features and contextual information (ION and MR-RCNN$\&$S-RCNN).

\hangafter 0
\hangindent 1em
\noindent
$\bullet$ It takes more time to train a more complex and deeper network (ResNet101 against VGG16) and this time consumption can be reduced by adding as many layers into shared fully convolutional layers as possible (FRCN).

\hangafter 0
\hangindent 1em
\noindent
$\bullet$ Regression based models can usually be processed in real-time at the cost of a drop in accuracy compared with region proposal based models. Also, region proposal based models can be modified into real-time systems with the introduction of other tricks \cite{pvanet} (PVANET), such as BN \cite{BN}, residual connections \cite{resnext}.
\section{Salient Object Detection}
Visual saliency detection, one of the most important and challenging tasks in computer vision, aims to highlight the most dominant object regions in an image. Numerous applications incorporate the visual saliency to improve their performance, such as image cropping \cite{rother2006autocollage} and segmentation \cite{jung2012unified}, image retrieval \cite{babenko2014neural} and object detection \cite{fpn}.

Broadly, there are two branches of approaches in salient object detection, namely bottom-up (BU) \cite{tu2016real} and top-down (TD) \cite{yang2017top}. Local feature contrast plays the central role in BU salient object detection, regardless of the semantic contents of the scene. To learn local feature contrast, various local and global features are extracted from pixels, e.g. edges \cite{rosin2009simple}, spatial information \cite{liu2011learning}. However, high-level and multi-scale semantic information cannot be explored with these low-level features. As a result, low contrast salient maps instead of salient objects are obtained. TD salient object detection is task-oriented and takes prior knowledge about object categories to guide the generation of salient maps. Taking semantic segmentation as an example, a saliency map is generated in the segmentation to assign pixels to particular object categories via a TD approach \cite{long2015fully}. In a word, TD saliency can be viewed as a focus-of-attention mechanism, which prunes BU salient points that are unlikely to be parts of the object \cite{gao2009discriminant}.
%Lee et al. [13] considered both high-level features extracted from CNNs and hand-crafted features. To combine them together, a unified fully connected neural network was designed to estimate saliency maps.
%Liu et al. [36] designed a two-stage deep network, in which a coarse prediction map was produced, followed by another network to refine the details of the prediction map hierarchically and progressively.
%Salient Object Subitizing %Unconstrained Salient Object Detection
\vspace{-0.3cm}
\subsection{Deep learning in Salient Object Detection}
Due to the significance for providing high-level and multi-scale feature representation and the successful applications in many correlated computer vision tasks, such as semantic segmentation \cite{long2015fully}, edge detection \cite{xie2015holistically} and generic object detection \cite{frcn}, it is feasible and necessary to extend CNN to salient object detection.

%Large-scale optimization of hierarchical features for saliency prediction in natural images
The early work by Eleonora Vig et al. \cite{vig2014large} follows a completely automatic data-driven approach to perform a large-scale search for optimal features, namely an ensemble of deep networks with different layers and parameters. To address the problem of limited training data, Kummerer et al. proposed the Deep Gaze \cite{kummerer2014deep} by transferring from the AlexNet to generate a high dimensional feature space and create a saliency map. A similar architecture was proposed by Huang et al. to integrate saliency prediction into pre-trained object recognition DNNs \cite{huang2015salicon}. The transfer is accomplished by fine-tuning DNNs' weights with an objective function based on the saliency evaluation metrics, such as Similarity, KL-Divergence and Normalized Scanpath Saliency.

Some works combined local and global visual clues to improve salient object detection performance.
%Deep networks for saliency detection via local estimation and global search
Wang et al. trained two independent deep CNNs (DNN-L and DNN-G) to capture local information and global contrast and predicted saliency maps by integrating both local estimation and global search \cite{LEGS}.
%Weakly Supervised Top-down Salient Object Detection
Cholakkal et al. proposed a weakly supervised saliency detection framework to combine visual saliency from bottom-up and top-down saliency maps, and refined the results with a multi-scale superpixel-averaging \cite{cholakkal2016weakly}.
%Saliency Detection by Multi-Context Deep Learning (seg+dec)
Zhao et al. proposed a multi-context deep learning framework, which utilizes a unified learning framework to model global and local context jointly with the aid of superpixel segmentation \cite{MC}.
%Two-Stream Convolutional Networks for Dynamic Saliency Prediction
To predict saliency in videos, Bak et al. fused two static saliency models, namely spatial stream net and temporal stream net, into a two-stream framework with a novel empirically grounded data augmentation technique \cite{bak2016two}.

Complementary information from semantic segmentation and context modeling is beneficial.
%SuperCNN: A Superpixelwise Convolutional Neural Network for Salient Object Detection (early paper)
To learn internal representations of saliency efficiently, He et al. proposed a novel superpixelwise CNN approach called SuperCNN \cite{supercnn}, in which salient object detection is formulated as a binary labeling problem. %DeepSaliency: Multi-Task Deep Neural Network Model for Salient Object Detection (figure 2,3) (seg+dec)
Based on a fully convolutional neural network, Li et al. proposed a multi-task deep saliency model, in which intrinsic correlations between saliency detection and semantic segmentation are set up \cite{MTDNN}. However, due to the conv layers with large receptive fields and pooling layers, blurry object boundaries and coarse saliency maps are produced.
%Saliency Detection via Combining Region-Level and Pixel-Level Predictions with CNNs
Tang et al. proposed a novel saliency detection framework (CRPSD) \cite{CRPSD}, which combines region-level saliency estimation and pixel-level saliency prediction together with three closely related CNNs.
%Deep contrast learning for salient object detection
Li et al. proposed a deep contrast network to combine segment-wise spatial pooling and pixel-level fully convolutional streams \cite{DCL}.

The proper integration of multi-scale feature maps is also of significance for improving detection performance.
%Edge Preserving and Multi-Scale Contextual Neural Network for Salient Object Detection
Based on Fast R-CNN, Wang et al. proposed the RegionNet by performing salient object detection with end-to-end edge preserving and multi-scale contextual modelling \cite{wang2016edge}. %Predicting Eye Fixations using Convolutional Neural Networks
Liu et al. \cite{liu2015predicting} proposed a multi-resolution convolutional neural network (Mr-CNN) to predict eye fixations, which is achieved by learning both bottom-up visual saliency and top-down visual factors from raw image data simultaneously. %A Deep Multi-Level Network for Saliency Prediction
Cornia et al. proposed an architecture which combines features extracted at different levels of the CNN \cite{cornia2016deep}. %Visual Saliency Detection Based on Multiscale Deep CNN Features
Li et al. proposed a multi-scale deep CNN framework to extract three scales of deep contrast features \cite{MDF}, namely the mean-subtracted region, the bounding box of its immediate neighboring regions and the masked entire image, from each candidate region.

It is efficient and accurate to train a direct pixel-wise CNN architecture to predict salient objects with the aids of RNNs and deconvolution networks. %Shallow and Deep Convolutional Networks for Saliency Prediction
Pan et al. formulated saliency prediction as a minimization optimization on the Euclidean distance between the predicted saliency map and the ground truth and proposed two kinds of architectures \cite{pan2016shallow}: a shallow one trained from scratch and a deeper one adapted from deconvoluted VGG network.
 %Recurrent Attentional Networks for Saliency Detection
As convolutional-deconvolution networks are not expert in recognizing objects of multiple scales, Kuen et al. proposed a recurrent attentional convolutional-deconvolution network (RACDNN) with several spatial transformer and recurrent network units to conquer this problem \cite{RACDNN}. %Deeply-Supervised Recurrent Convolutional Neural Network for Saliency Detection
To fuse local, global and contextual information of salient objects, Tang et al. developed a deeply-supervised recurrent convolutional neural network (DSRCNN) to perform a full image-to-image saliency detection \cite{DSRCNN}.
\vspace{-0.2cm}
\subsection{Experimental Evaluation}
\begin{table*}[!tb]
\caption{Comparison between state of the art methods.}
\vspace{-0.2cm}
\setlength{\tabcolsep}{4pt}
\renewcommand\arraystretch{0.7}
\label{tab:salient}
\centering
\tiny
\begin{threeparttable}[b]
\begin{tabular}{c|c|c|c|c|c|c|c|c|c|c|c|c|c|c}
\toprule[2pt]
Dataset	&Metrics	&CHM\cite{CHM}	&RC\cite{RC}	&DRFI\cite{DRFI}	&MC\cite{MC} &MDF\cite{MDF}	&LEGS\cite{LEGS}	&DSR\cite{DSRCNN} &MTDNN\cite{MTDNN} &CRPSD\cite{CRPSD}	&DCL\cite{DCL}	&ELD\cite{ELD}	&NLDF\cite{NLDF}&DSSC\cite{DSSC}  \tabularnewline
\midrule[2pt]
\multirow{2}{*}{PASCAL-S}&$wF_{\beta}$	&0.631	&0.640	&0.679	&0.721	&0.764	&0.756	&0.697	&0.818	&0.776	&0.822	&0.767	&\bf 0.831	& 0.830 \tabularnewline
&MAE	&0.222	&0.225	&0.221	&0.147	&0.145	&0.157	&0.128	&0.170	& \bf 0.063	&0.108	&0.121	&0.099	&  0.080 \tabularnewline
\midrule[2pt]
\multirow{2}{*}{ECSSD}&$wF_{\beta}$	&0.722	&0.741	&0.787	&0.822	&0.833	&0.827	&0.872	&0.810	&0.849	&0.898	&0.865	&0.905	&\bf 0.915 \tabularnewline
&MAE &0.195	&0.187	&0.166	&0.107	&0.108	&0.118	&\bf 0.037	&0.160	&0.046	&0.071	&0.098	&0.063	&0.052 \tabularnewline
\midrule[2pt]
\multirow{2}{*}{HKU-IS}&$wF_{\beta}$ &0.728	&0.726	&0.783	&0.781	&0.860	&0.770	&0.833	&-	&0.821	&0.907	&0.844	&0.902	&\bf 0.913 \tabularnewline
&MAE &0.158	&0.165	&0.143	&0.098	&0.129	&0.118	&0.040	&-	&0.043	&0.048	&0.071	&0.048	&\bf 0.039 \tabularnewline
\midrule[2pt]
\multirow{2}{*}{SOD}&$wF_{\beta}$ &0.655	&0.657	&0.712	&0.708	&0.785	&0.707	&-	&0.781	&-	&0.832	&0.760	&0.810	&\bf 0.842\tabularnewline
&MAE &0.249	&0.242	&0.215	&0.184	&0.155	&0.205	&-	&0.150	&-	&0.126	&0.154	&0.143	&\bf 0.118 \tabularnewline
\bottomrule[2pt]
\end{tabular}
\begin{tablenotes}
  \item[*] The bigger $wF_{\beta}$ is or the smaller MAE is, the better the performance is.
\end{tablenotes}
\end{threeparttable}
\vspace{-0.7cm}
\end{table*}

Four representative datasets, including ECSSD \cite{ECSSD}, HKU-IS \cite{MDF}, PASCALS \cite{PASCALS}, and SOD \cite{SOD}, are used to evaluate several state-of-the-art methods. ECSSD consists of 1000 structurally complex but semantically meaningful natural images. HKU-IS is a large-scale dataset containing over 4000 challenging images. Most of these images have more than one salient object and own low contrast. PASCALS is a subset chosen from the validation set of PASCAL VOC 2010 segmentation dataset and is composed of 850 natural images. The SOD dataset possesses 300 images containing multiple salient objects. The training and validation sets for different datasets are kept the same as those in \cite{DRFI}.

Two standard metrics, namely F-measure and the mean absolute error (MAE), are utilized to evaluate the quality of a saliency map. Given precision and recall values pre-computed on the union of generated binary mask $B$ and ground truth $Z$, F-measure is defined as below
\begin{equation}
F_{\beta}=\frac{(1+\beta^2)Presion\times Recall}{\beta^2Presion+ Recall}
\end{equation}
where $\beta^2$ is set to 0.3 in order to stress the importance of the precision value.

The MAE score is computed with the following equation
\begin{equation}
MAE=\frac{1}{H\times W}\sum_{i=1}^{H}\sum_{j=1}^{W}\left|\hat{S}(i,j)=\hat{Z}(i,j)\right|
\end{equation}
where $\hat{Z}$ and $\hat{S}$ represent the ground truth and the continuous saliency map, respectively. $W$ and $H$ are the width and height of the salient area, respectively. This score stresses the importance of successfully detected salient objects over detected non-salient pixels \cite{salient-benchmark}.

The following approaches are evaluated: CHM \cite{CHM}, RC \cite{RC}, DRFI \cite{DRFI}, MC \cite{MC}, MDF \cite{MDF}, LEGS \cite{LEGS}, DSR \cite{DSRCNN}, MTDNN \cite{MTDNN}, CRPSD \cite{CRPSD}, DCL \cite{DCL}, ELD \cite{ELD}, NLDF \cite{NLDF} and DSSC \cite{DSSC}. Among these methods, CHM, RC and DRFI are classical ones with the best performance \cite{salient-benchmark}, while the other methods are all associated with CNN. F-measure and MAE scores are shown in Table \ref{tab:salient}.

From Table \ref{tab:salient}, we can find that CNN based methods perform better than classic methods. MC and MDF combine the information from local and global context to reach a more accurate saliency. ELD refers to low-level handcrafted features for complementary information. LEGS adopts generic region proposals to provide initial salient regions, which may be insufficient for salient detection. DSR and MT act in different ways by introducing recurrent network and semantic segmentation, which provide insights for future improvements. CPRSD, DCL, NLDF and DSSC are all based on multi-scale representations and superpixel segmentation, which provide robust salient regions and smooth boundaries. DCL, NLDF and DSSC perform the best on these four datasets. DSSC earns the best performance by modelling scale-to-scale short-connections.

Overall, as CNN mainly provides salient information in local regions, most of CNN based methods need to model visual saliency along region boundaries with the aid of superpixel segmentation. Meanwhile, the extraction of multi-scale deep CNN features is of significance for measuring local conspicuity. Finally, it's necessary to strengthen local connections between different CNN layers and as well to utilize complementary information from local and global context.
\section{Face Detection}
Face detection is essential to many face applications and acts as an important pre-processing procedure to face recognition \cite{Peng2015Graphical,Peng2016Face,Gao2012Face}, face synthesis \cite{Wang2014A,Peng2016Multiple} and facial expression analysis \cite{Majumder2018AutomaticFE}. Different from generic object detection, this task is to recognize and locate face regions covering a very large range of scales (30-300 pts vs. 10-1000 pts). At the same time, faces have their unique object structural configurations (e.g. the distribution of different face parts) and characteristics (e.g. skin color). All these differences lead to special attention to this task. However, large visual variations of faces, such as occlusions, pose variations and illumination changes, impose great challenges for this task in real applications.

The most famous face detector proposed by Viola and Jones \cite{Viola2004Robust} trains cascaded classifiers with Haar-Like features and AdaBoost, achieving good performance with real-time efficiency. However, this detector may degrade significantly in real-world applications due to larger visual variations of human faces. Different from this cascade structure, Felzenszwalb et al. proposed a deformable part model (DPM) for face detection \cite{DPM}. However, for these traditional face detection methods, high computational expenses and large quantities of annotations are required to achieve a reasonable result. Besides, their performance is greatly restricted by manually designed features and shallow architecture.
\vspace{-0.3cm}
\subsection{Deep learning in Face Detection}
Recently, some CNN based face detection approaches have been proposed \cite{unitbox,DDFD,faceness}.%UnitBox: An Advanced Object Detection Network (IOU loss)
As less accurate localization results from independent regressions of object coordinates, Yu et al. \cite{unitbox} proposed a novel IoU loss function for predicting the four bounds of box jointly.
%Multi-view Face Detection Using Deep Convolutional Neural Networks (slide windows+simpfied R-CNN)
%While there has been significant research on this problem, current state-of-the-art approaches for this task require annotation of facial landmarks, e.g. TSM [25], or annotation of face poses [28, 22]. They also require training dozens of models to fully capture faces in all orientations, e.g. 22 models in HeadHunter method [22]. [9]
Farfade et al. \cite{DDFD} proposed a Deep Dense Face Detector (DDFD) to conduct multi-view face detection, which is able to detect faces in a wide range of orientations without requirement of pose/landmark annotations.
%From Facial Parts Responses to Face Detection: A Deep Learning Approach [23]
Yang et al. proposed a novel deep learning based face detection framework \cite{faceness}, which collects the responses from local facial parts (e.g. eyes, nose and mouths) to address face detection under severe occlusions and unconstrained pose variations. Yang et al. \cite{ScaleFace} proposed a scale-friendly detection network named ScaleFace, which splits a large range of target scales into smaller sub-ranges. Different specialized sub-networks are constructed on these sub-scales and combined into a single one to conduct end-to-end optimization. Hao et al. designed an efficient CNN to predict the scale distribution histogram of the faces and took this histogram to guide the zoom-in and zoom-out of the image \cite{Hao2017ScaleAwareFD}. Since the faces are approximately in uniform scale after zoom, compared with other state-of-the-art baselines, better performance is achieved with less computation cost. Besides, some generic detection frameworks are extended to face detection with different modifications, e.g. Faster R-CNN \cite{face-rcnn,DeepIR,faster_face}.

Some authors trained CNNs with other complementary tasks, such as 3D modelling and face landmarks, in a multi-task learning manner.
%DenseBox: Unifying Landmark Localization with End to End Object Detection
Huang et al. proposed a unified end-to-end FCN framework called DenseBox to jointly conduct face detection and landmark localization \cite{densebox}.
%Face Detection with End-to-End Integration of a ConvNet and a 3D Model (multi-task learning with bounding box regression, 3D face modelling and classficiation, bold one) the FDDB benchmark [19] with fine-tuning and on the AFW benchmark [41]
Li et al. \cite{conv3d} proposed a multi-task discriminative learning framework which integrates a ConvNet with a fixed 3D mean face model in an end-to-end manner. In the framework, two issues are addressed to transfer from generic object detection to face detection, namely eliminating predefined anchor boxes by a 3D mean face model and replacing RoI pooling layer with a configuration pooling layer.
%MTCNN: Joint Face Detection and Alignment using Multi-task Cascaded Convolutional Networks
Zhang et al. \cite{MTCNN} proposed a deep cascaded multi-task framework named MTCNN which exploits the inherent correlations between face detection and alignment in unconstrained environment to boost up detection performance in a coarse-to-fine manner.

Reducing computational expenses is of necessity in real applications.
%Compact Convolutional Neural Network Cascade for Face Detection
To achieve real-time detection on mobile platform, Kalinovskii and Spitsyn proposed a new solution of frontal face detection based on compact CNN cascades \cite{kalinovsky2016compact}. This method takes a cascade of three simple CNNs to generate, classify and refine candidate object positions progressively.
%Supervised Transformer Network for Efficient Face Detection
To reduce the effects of large pose variations, Chen et al. proposed a cascaded CNN denoted by Supervised Transformer Network \cite{chen2016supervised}. This network takes a multi-task RPN to predict candidate face regions along with associated facial landmarks simultaneously, and adopts a generic R-CNN to verify the existence of valid faces.
%A Multi-Scale Cascade Fully Convolutional Network Face Detector
Yang et al. proposed a three-stage cascade structure based on FCNs \cite{yang2016multi}, while in each stage, a multi-scale FCN is utilized to refine the positions of possible faces. Qin et al. proposed a unified framework which achieves better results with the complementary information from different jointly trained CNNs \cite{Qin2016Joint}.
\vspace{-0.3cm}
\subsection{Experimental Evaluation}
The FDDB \cite{fddb} dataset has a total of 2,845 pictures in which 5,171 faces are annotated with elliptical shape. Two types of evaluations are used: the discrete score and continuous score. By varying the threshold of the decision rule, the ROC curve for the discrete scores can reflect the dependence of the detected face fractions on the number of false alarms. Compared with annotations, any detection with an IoU ratio exceeding 0.5 is treated as positive. Each annotation is only associated with one detection. The ROC curve for the continuous scores is the reflection of face localization quality.

\begin{figure}[!tb]
    \centering
    \subfloat[Discrete ROC curves]{
    \label{fig:dis_fddb}
    \includegraphics[width=0.4\textwidth, height= 3.6cm]{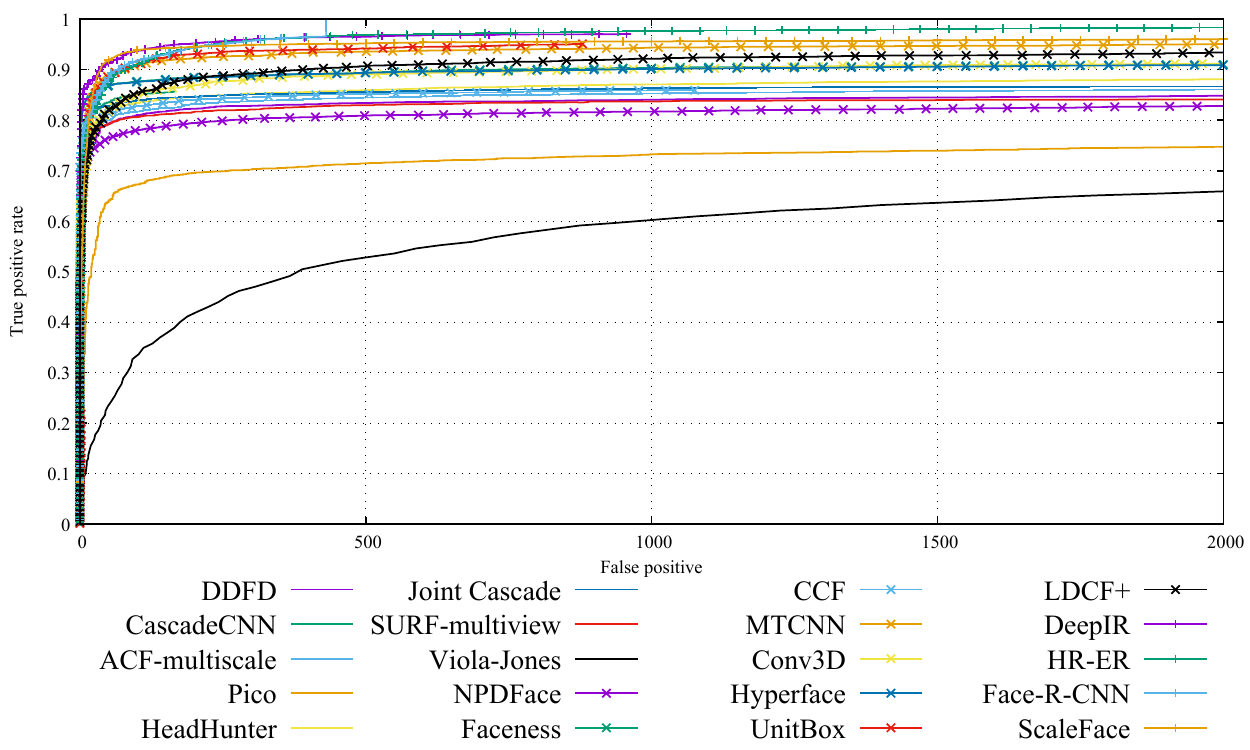}
    }
   \vfill
\vspace{-0.3cm}
    \subfloat[Continuous ROC curves]{
    \label{fig:con_fddb}
    \includegraphics[width=0.4\textwidth, height= 3.6cm]{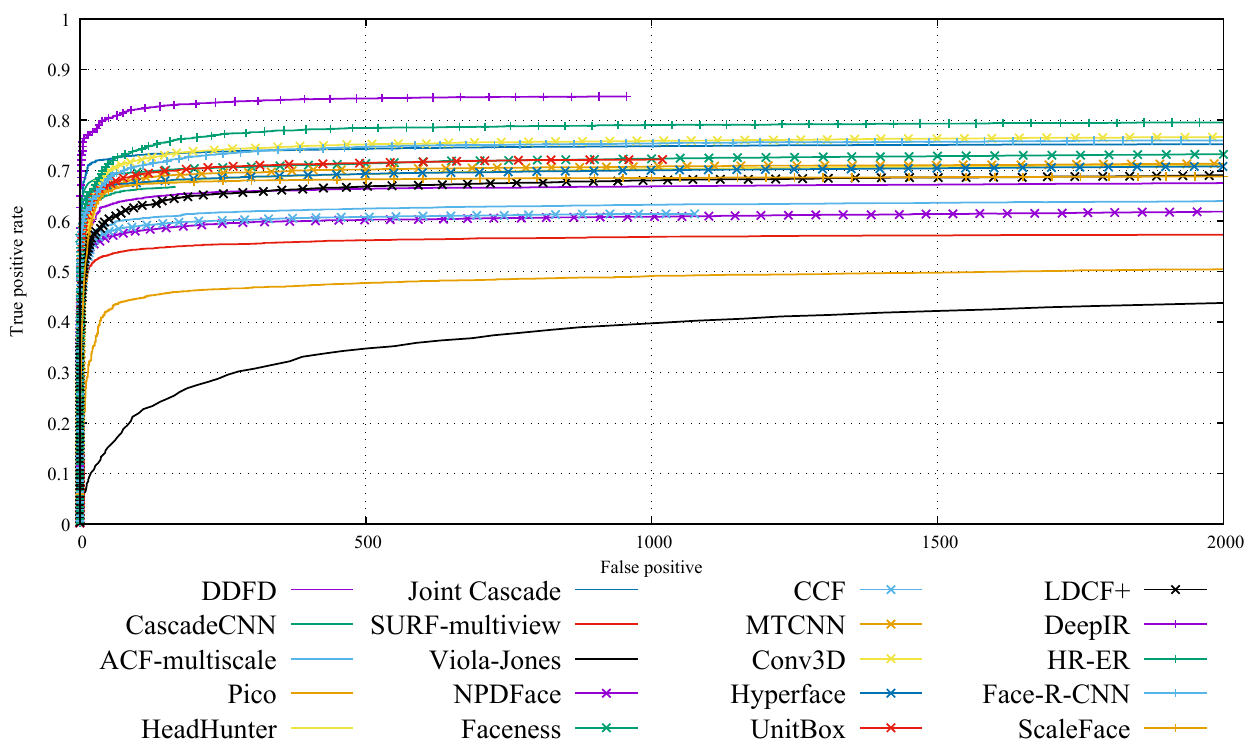}
    }
    \caption{The ROC curves of state-of-the-art methods on FDDB.}
     \label{fig:fddb}
\vspace{-0.5cm}
    \end{figure}

The evaluated models cover DDFD \cite{DDFD}, CascadeCNN \cite{CascadeCNN}, ACF-multiscale \cite{ACF-multiscale}, Pico \cite{Pico}, HeadHunter \cite{HeadHunter}, Joint Cascade \cite{Joint-cascade}, SURF-multiview \cite{SURF-multiview}, Viola-Jones \cite{Viola2004Robust}, NPDFace \cite{NPDFace}, Faceness \cite{faceness}, CCF \cite{CCF}, MTCNN \cite{MTCNN}, Conv3D \cite{conv3d}, Hyperface \cite{hyperface}, UnitBox \cite{unitbox}, LDCF+ [S2], DeepIR \cite{DeepIR}, HR-ER \cite{HR-ER}, Face-R-CNN \cite{face-rcnn} and ScaleFace \cite{ScaleFace}. ACF-multiscale, Pico, HeadHunter, Joint Cascade, SURF-multiview, Viola-Jones, NPDFace and LDCF+ are built on classic hand-crafted features while the rest methods are based on deep CNN features. The ROC curves are shown in Figure \ref{fig:fddb}.

From Figure \ref{fig:fddb}(a), in spite of relatively competitive results produced by LDCF+, it can be observed that most of classic methods perform with similar results and are outperformed by CNN based methods by a significant margin. From Figure \ref{fig:fddb}(b), it can be observed that most of CNN based methods earn similar true positive rates between 60\% and 70\% while DeepIR and HR-ER perform much better than them. Among classic methods, Joint Cascade is still competitive. As earlier works, DDFD and CCF directly make use of generated feature maps and obtain relatively poor results. CascadeCNN builds cascaded CNNs to locate face regions, which is efficient but inaccurate. Faceness combines the decisions from different part detectors, resulting in precise face localizations while being time-consuming. The outstanding performance of MTCNN, Conv3D and Hyperface proves the effectiveness of multi-task learning. HR-ER and ScaleFace adaptively detect faces of different scales, and make a balance between accuracy and efficiency. DeepIR and Face-R-CNN are two extensions of the Faster R-CNN architecture to face detection, which validate the significance and effectiveness of Faster R-CNN. Unitbox provides an alternative choice for performance improvements by carefully designing optimization loss.

From these results, we can draw the conclusion that CNN based methods are in the leading position. The performance can be improved by the following strategies: designing novel optimization loss, modifying generic detection pipelines, building meaningful network cascades, adapting scale-aware detection and learning multi-task shared CNN features.
\section{Pedestrian Detection}
Recently, pedestrian detection has been intensively studied, which has a close relationship to pedestrian tracking \cite{Jiang2018MultiplePT,Gavrila2006MulticuePD}, person re-identification \cite{Xu2017JointlyAS,Liu2017StepwiseMP} and robot navigation \cite{Khan2018CooperativeRT,Geiger2013Vision}. Prior to the recent progress in DCNN based methods \cite{CompACT-Deep,tian2015deep}, some researchers combined boosted decision forests with hand-crafted features to obtain pedestrian detectors \cite{ACF,Checkerboard+,pedestrian_optical}. At the same time, to explicitly model the deformation and occlusion, part-based models \cite{lin2015discriminatively} and explicit occlusion handling \cite{mathias2013handling,tang2014detection} are of concern. %% delete references

As there are many pedestrian instances of small sizes in typical scenarios of pedestrian detection (e.g. automatic driving and intelligent surveillance), the application of RoI pooling layer in generic object detection pipeline may result in `plain' features due to collapsing bins. In the meantime, the main source of false predictions in pedestrian detection is the confusion of hard background instances, which is in contrast to the interference from multiple categories in generic object detection. As a result, different configurations and components are required to accomplish accurate pedestrian detection.
\vspace{-0.3cm}
\subsection{Deep learning in Pedestrian Detection}
Although DCNNs have obtained excellent performance on generic object detection \cite{frcn,yolov2}, none of these approaches have achieved better results than the best hand-crafted feature based method \cite{Checkerboard+} for a long time, even when part-based information and occlusion handling are incorporated \cite{tang2014detection}. Thereby, some researches have been conducted to analyze the reasons.
%Is Faster R-CNN Doing Well for Pedestrian Detection?
Zhang et al. attempted to adapt generic Faster R-CNN \cite{Faster} to pedestrian detection \cite{rpn+bf}. They modified the downstream classifier by adding boosted forests to shared, high-resolution conv feature maps and taking a RPN to handle small instances and hard negative examples. %Deep Learning Strong Parts for Pedestrian Detection (DPM+CNN)
To deal with complex occlusions existing in pedestrian images, inspired by DPM \cite{DPM}, Tian et al. proposed a deep learning framework called DeepParts \cite{deepparts}, which makes decisions based an ensemble of extensive part detectors. DeepParts has advantages in dealing with weakly labeled data, low IoU positive proposals and partial occlusion.

Other researchers also tried to combine complementary information from multiple data sources. CompACT-Deep adopts a complexity-aware cascade to combine hand-crafted features and fine-tuned DCNNs \cite{CompACT-Deep}. %Multispectral Deep Neural Networks for Pedestrian Detection
Based on Faster R-CNN, Liu et al. proposed multi-spectral deep neural networks for pedestrian detection to combine complementary information from color and thermal images \cite{liu2016multispectral}.
%Pedestrian Detection aided by Deep Learning Semantic Tasks(TACNN£¬2014, too complex and must bias all the tasks carefully(pedestrian classifier and attributes, shared and unshared bkg attributes))
Tian et al. \cite{TACNN} proposed a task-assistant CNN (TA-CNN) to jointly learn multiple tasks with multiple data sources and to combine pedestrian attributes with semantic scene attributes together. %Fused DNN: A deep neural network fusion approach to fast and robust pedestrian detection
Du et al. proposed a deep neural network fusion architecture for fast and robust pedestrian detection \cite{f-dnn}. Based on the candidate bounding boxes generated with SSD detectors \cite{ssd}, multiple binary classifiers are processed parallelly to conduct soft-rejection based network fusion (SNF) by consulting their aggregated degree of confidences.

However, most of these approaches are much more sophisticated than the standard R-CNN framework. CompACT-Deep consists of a variety of hand-crafted features, a small CNN model and a large VGG16 model \cite{CompACT-Deep}. DeepParts contains 45 fine-tuned DCNN models, and a set of strategies, including bounding box shifting handling and part selection, are required to arrive at the reported results \cite{deepparts}. So the modification and simplification is of significance to reduce the burden on both software and hardware to satisfy real-time detection demand. %Deep convolutional neural networks for pedestrian detection
Tome et al. proposed a novel solution to adapt generic object detection pipeline to pedestrian detection by optimizing most of its stages \cite{tome2016deep}. %Pushing the Limits of Deep CNNs for Pedestrian Detection
Hu et al. \cite{hu2017pushing} trained an ensemble of boosted decision models by reusing the conv feature maps, and a further improvement was gained with simple pixel labelling and additional complementary hand-crafted features. %Reduced Memory Region Based Deep Convolutional Neural Network Detection (reduce cost)
Tome et al. \cite{tome2016reduced} proposed a reduced memory region based deep CNN architecture, which fuses regional responses from both ACF detectors and SVM classifiers into R-CNN.
%A Real-Time Pedestrian Detector using Deep Learning for Human-Aware Navigation
Ribeiro et al. addressed the problem of Human-Aware Navigation \cite{ribeiro2016real} and proposed a vision-based person tracking system guided by multiple camera sensors.
\subsection{Experimental Evaluation}
\begin{table}[!tb]
\caption{Detailed breakdown performance comparisons of state-of-the-art models on Caltech Pedestrian dataset. All numbers are reported in L-AMR. }
\vspace{-0.2cm}
\renewcommand\arraystretch{0.9}
\setlength{\tabcolsep}{4pt}
\label{tab:caltech}
\centering
\tiny
\begin{tabular}{c|c|c|c|c|c|c|c|c}
%{p{1.3cm}<{\centering}p{1.3cm}<{\centering}p{1.3cm}<{\centering}p{1.3cm}<{\centering}p{1.3cm}<{\centering}p{1.3cm}<{\centering}p{1.3cm}<{\centering}p{1.3cm}<{\centering}p{1.3cm}<{\centering}}
\toprule[2pt]
Method	&Reasonable	&All	&Far	&Medium	&Near	& none	&partial	&heavy \tabularnewline
 \midrule[2pt]
Checkerboards+\cite{Checkerboard+}	&17.1	&68.4	&100	&58.3	&5.1	&15.6	&31.4	&78.4\tabularnewline
LDCF++[S2]	&15.2	&67.1	&100	&58.4	&5.4	&13.3	&33.3	&76.2\tabularnewline
 SCF+AlexNet\cite{SCF+AlexNet}	&23.3	&70.3	&100	&62.3	&10.2	&20.0	&48.5	&74.7\tabularnewline
SA-FastRCNN\cite{SA-FastRCNN}	&9.7	&62.6	&100	&51.8	&\bf 0	&7.7	&24.8	&64.3\tabularnewline
MS-CNN\cite{mscnn}	&10.0	&61.0	&97.2	&49.1	&2.6	&8.2	&19.2	&60.0 \tabularnewline
DeepParts\cite{deepparts}	&11.9	&64.8	&100	&56.4	&4.8	&10.6	&19.9	&60.4\tabularnewline
CompACT-Deep\cite{CompACT-Deep}	&11.8	&64.4	&100	&53.2	&4.0	&9.6	&25.1	&65.8\tabularnewline
RPN+BF\cite{rpn+bf}	&9.6	&64.7	&100	&53.9	&2.3	&7.7	&24.2	&74.2\tabularnewline
F-DNN+SS\cite{f-dnn}	&\bf 8.2	&\bf 50.3	&\bf 77.5	&\bf 33.2	&2.8	&\bf 6.7	&\bf 15.1	&\bf 53.4\tabularnewline
\bottomrule[2pt]
\end{tabular}
\par
\vspace{-0.5cm}
\end{table}
The evaluation is conducted on the most popular Caltech Pedestrian dataset \cite{Wojek2012Pedestrian}. The dataset was collected from the videos of a vehicle driving through an urban environment and consists of 250,000 frames with about 2300 unique pedestrians and 350,000 annotated bounding boxes (BBs). Three kinds of labels, namely `Person (clear identifications)', `Person? (unclear identifications)' and `People (large group of individuals)', are assigned to different BBs. The performance is measured with the log-average miss rate (L-AMR) which is computed evenly spaced in log-space in the range $10^{-2}$ to 1 by averaging miss rate at the rate of nine false positives per image (FPPI) \cite{Wojek2012Pedestrian}. According to the differences in the height and visible part of the BBs, a total of 9 popular settings are adopted to evaluate different properties of these models. Details of these settings are as \cite{Wojek2012Pedestrian}.

Evaluated methods include Checkerboards+ \cite{Checkerboard+}, LDCF++ [S2], SCF+AlexNet \cite{SCF+AlexNet}, SA-FastRCNN \cite{SA-FastRCNN}, MS-CNN \cite{mscnn}, DeepParts \cite{deepparts}, CompACT-Deep \cite{CompACT-Deep}, RPN+BF \cite{rpn+bf} and F-DNN+SS \cite{f-dnn}. The first two methods are based on hand-crafted features while the rest ones rely on deep CNN features. All results are exhibited in Table \ref{tab:caltech}. From this table, we observe that different from other tasks, classic handcrafted features can still earn competitive results with boosted decision forests \cite{rpn+bf}, ACF \cite{ACF} and HOG+LUV channels [S2]. As an early attempt to adapt CNN to pedestrian detection, the features generated by SCF+AlexNet are not so discriminant and produce relatively poor results. Based on multiple CNNs, DeepParts and CompACT-Deep accomplish detection tasks via different strategies, namely local part integration and cascade network. The responses from different local part detectors make DeepParts robust to partial occlusions. However, due to complexity, it is too time-consuming to achieve real-time detection. The multi-scale representation of MS-CNN improves accuracy of pedestrian locations. SA-FastRCNN extends Fast R-CNN to automatically detecting pedestrians according to their different scales, which has trouble when there are partial occlusions. RPN+BF combines the detectors produced by Faster R-CNN with boosting decision forest to accurately locate different pedestrians. F-DNN+SS, which is composed of multiple parallel classifiers with soft rejections, performs the best followed by RPN+BF, SA-FastRCNN and MS-CNN.

In short, CNN based methods can provide more accurate candidate boxes and multi-level semantic information for identifying and locating pedestrians. Meanwhile, handcrafted features are complementary and can be combined with CNN to achieve better results. The improvements over existing CNN methods can be obtained by carefully designing the framework and classifiers, extracting multi-scale and part based semantic information and searching for complementary information from other related tasks, such as segmentation.
\section{Promising Future Directions and Tasks}
In spite of rapid development and achieved promising progress of object detection, there are still many open issues for future work.

The first one is small object detection such as occurring in COCO dataset and in face detection task. To improve localization accuracy on small objects under partial occlusions, it is necessary to modify network architectures from the following aspects.

\hangafter 0
\hangindent 1em
\noindent
$\bullet$ \textbf{\emph{Multi-task joint optimization and multi-modal information fusion.}} Due to the correlations between different tasks within and outside object detection, multi-task joint optimization has already been studied by many researchers \cite{frcn}\cite{Faster}. However, apart from the tasks mentioned in Subs. \ref{sec:m_m_c}, it is desirable to think over the characteristics of different sub-tasks of object detection (e.g. superpixel semantic segmentation in salient object detection) and extend multi-task optimization to other applications such as instance segmentation \cite{fpn}, multi-object tracking \cite{tang2014detection} and multi-person pose estimation [S4]. Besides, given a specific application, the information from different modalities, such as text \cite{gao2013visual}, thermal data \cite{liu2016multispectral} and images \cite{rfcn}, can be fused together to achieve a more discriminant network.

\hangafter 0
\hangindent 1em
\noindent
$\bullet$ \textbf{\emph{Scale adaption.}} Objects usually exist in different scales, which is more apparent in face detection and pedestrian detection. To increase the robustness to scale changes, it is demanded to train scale-invariant, multi-scale or scale-adaptive detectors. For scale-invariant detectors, more powerful backbone architectures (e.g. ResNext \cite{resnext}), negative sample mining \cite{OHEM}, reverse connection \cite{RONRC} and sub-category modelling \cite{subcnn} are all beneficial. For multi-scale detectors, both the FPN \cite{fpn} which produces multi-scale feature maps and Generative Adversarial Network \cite{GAN} which narrows representation differences between small objects and the large ones with a low-cost architecture provide insights into generating meaningful feature pyramid. For scale-adaptive detectors, it is useful to combine knowledge graph \cite{Fang2017ObjectDM}, attentional mechanism \cite{Welleck2017SaliencybasedSI}, cascade network \cite{CascadeCNN} and scale distribution estimation \cite{Hao2017ScaleAwareFD} to detect objects adaptively.

\hangafter 0
\hangindent 1em
\noindent
$\bullet$ \textbf{\emph{Spatial correlations and contextual modelling.}} Spatial distribution plays an important role in object detection. So region proposal generation and grid regression are taken to obtain probable object locations. However, the correlations between multiple proposals and object categories are ignored. Besides, the global structure information is abandoned by the position-sensitive score maps in R-FCN. To solve these problems, we can refer to diverse subset selection \cite{DPP} and sequential reasoning tasks \cite{Sukhbaatar2015EndToEndMN} for possible solutions. It is also meaningful to mask salient parts and couple them with the global structure in a joint-learning manner \cite{Dabkowski2017Real}.

The second one is to release the burden on manual labor and accomplish real-time object detection, with the emergence of large-scale image and video data. The following three aspects can be taken into account.

\hangafter 0
\hangindent 1em
\noindent
$\bullet$ \textbf{\emph{Cascade network.}} In a cascade network, a cascade of detectors are built in different stages or layers \cite{craft,CascadeCNN}. And easily distinguishable examples are rejected at shallow layers so that features and classifiers at latter stages can handle more difficult samples with the aid of the decisions from previous stages. However, current cascades are built in a greedy manner, where previous stages in cascade are fixed when training a new stage. So the optimizations of different CNNs are isolated, which stresses the necessity of end-to-end optimization for CNN cascade. At the same time, it is also a matter of concern to build contextual associated cascade networks with existing layers.

\hangafter 0
\hangindent 1em
\noindent
$\bullet$ \textbf{\emph{Unsupervised and weakly supervised learning.}} It's very time consuming to manually draw large quantities of bounding boxes. To release this burden, semantic prior \cite{noh2015learning}, unsupervised object discovery \cite{Croitoru2017UnsupervisedLF}, multiple instance learning \cite{Wang2014WeaklySO} and deep neural network prediction \cite{resnet} can be integrated to make best use of image-level supervision to assign object category tags to corresponding object regions and refine object boundaries. Furthermore, weakly annotations (e.g. center-click annotations \cite{Papadopoulos2017TrainingOC}) are also helpful for achieving high-quality detectors with modest annotation efforts, especially aided by the mobile platform.

\hangafter 0
\hangindent 1em
\noindent
$\bullet$ \textbf{\emph{Network optimization.}} Given specific applications and platforms, it is significant to make a balance among speed, memory and accuracy by selecting \textbf{an} optimal detection architecture \cite{pvanet,Huang2017SpeedAccuracyTF}. However, despite that detection accuracy is reduced, it is more meaningful to learn compact models with fewer number of parameters \cite{tome2016reduced}. And this situation can be relieved by introducing better pre-training schemes \cite{Li2017MimickingVE}, knowledge distillation \cite{Hinton2015Distilling} and hint learning \cite{Romero2014FitNets}. DSOD also provides a promising guideline to train from scratch to bridge the gap between different image sources and tasks \cite{DSOD}.

The third one is to extend typical methods for 2D object detection to adapt 3D object detection and video object detection, with the requirements from autonomous driving, intelligent transportation and intelligent surveillance.

\hangafter 0
\hangindent 1em
\noindent
$\bullet$ \textbf{\emph{3D object detection}.} With the applications of 3D sensors (e.g. LIDAR and camera), additional depth information can be utilized to better understand the images in 2D and extend the image-level knowledge to the real world. However, seldom of these 3D-aware techniques aim to place correct 3D bounding boxes around detected objects. To achieve better bounding results, multi-view representation \cite{ACF-multiscale} and 3D proposal network \cite{DOP} may provide some guidelines to encode depth information with the aid of inertial sensors (accelerometer and gyrometer) \cite{Dong2017VisualInertialSemanticSR}.

\hangafter 0
\hangindent 1em
\noindent
$\bullet$ \textbf{\emph{Video object detection.}} Temporal information across different frames play an important role in understanding the behaviors of different objects. However, the accuracy suffers from degenerated object appearances (e.g., motion blur and video defocus) in videos and the network is usually not trained end-to-end. To this end, spatiotemporal tubelets \cite{Kang2017ObjectDI}, optical flow \cite{pedestrian_optical} and LSTM \cite{lstm_scene} should be considered to fundamentally model object associations between consecutive frames.
\section{Conclusion}
Due to its powerful learning ability and advantages in dealing with occlusion, scale transformation and background switches, deep learning based object detection has been a research hotspot in recent years. This paper provides a detailed review on deep learning based object detection frameworks which handle different sub-problems, such as occlusion, clutter and low resolution, with different degrees of modifications on R-CNN. The review starts on generic object detection pipelines which provide base architectures for other related tasks. Then, three other common tasks, namely salient object detection, face detection and pedestrian detection, are also briefly reviewed. Finally, we propose several promising future directions to gain a thorough understanding of the object detection landscape. This review is also meaningful for the developments in neural networks and related learning systems, which provides valuable insights and guidelines for future progress.

\section*{Acknowledgments}
This research was supported by the National Natural Science Foundation of China (No.61672203 $\&$ 61375047 $\&$ 91746209), the National Key Research and Development Program of China (2016YFB1000901), and Anhui Natural Science Funds for Distinguished Young Scholar (No.170808J08).
\ifCLASSOPTIONcaptionsoff
  \newpage
\fi



\bibliographystyle{IEEEtran}
\bibliography{IEEEabrv,IEEErefs-proof}
%\begin{thebibliography}{1}
%\end{thebibliography}

\begin{IEEEbiography}[{\includegraphics[width=1in,height=1.25in,clip,keepaspectratio]{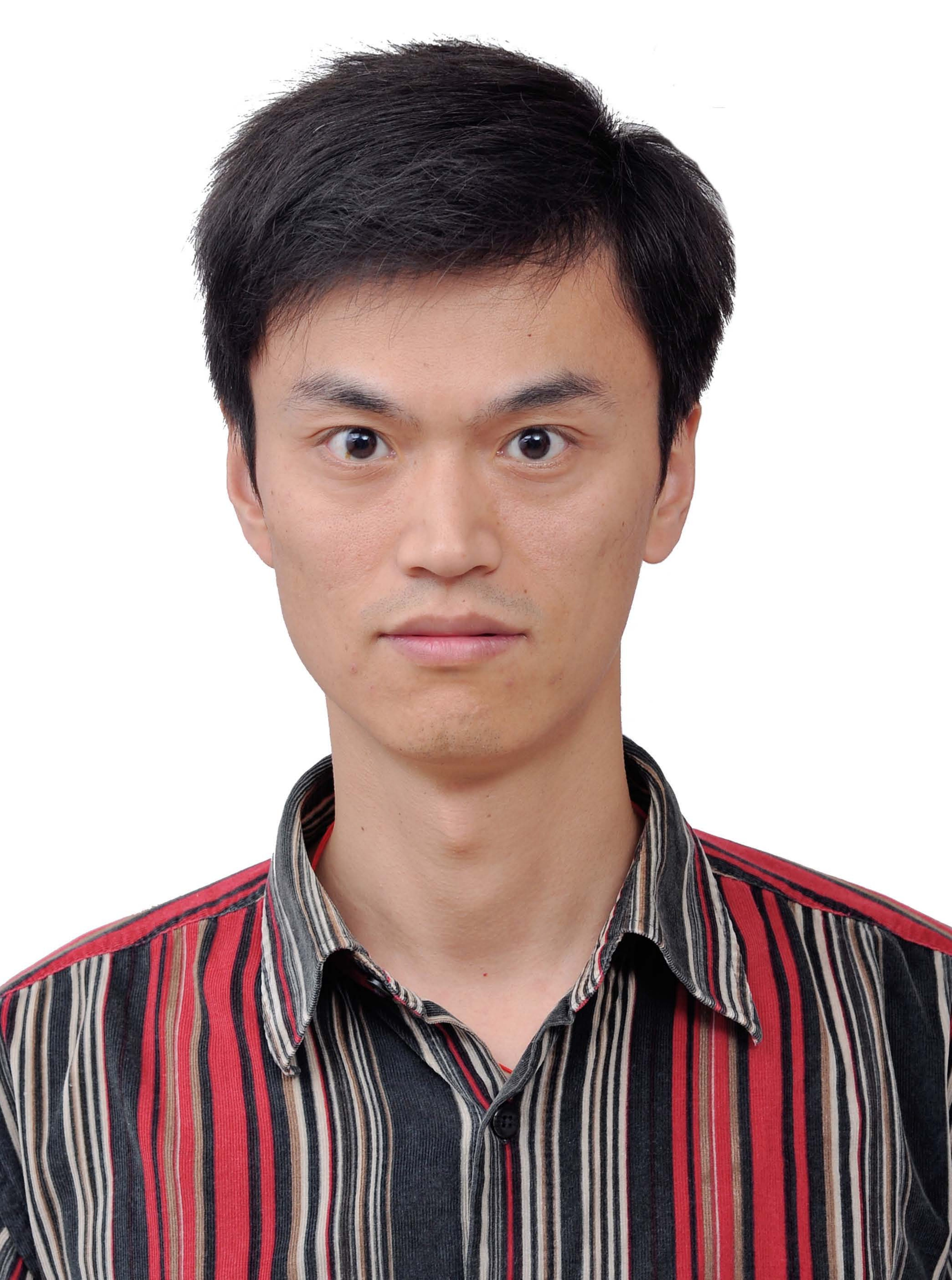}}]{Zhong-Qiu Zhao}
is a professor at Hefei University of Technology, China. He obtained the Ph.D. degree in Pattern Recognition \& Intelligent System at University of Science and Technology, China, in 2007. From April 2008 to November 2009, he held a postdoctoral position in image processing in CNRS UMR6168 Lab Sciences de l¡¯Information et des Syst\`emes, France. From January 2013 to December 2014, he held a research fellow position in image processing at the Department of Computer Science of Hongkong Baptist University, Hongkong, China. His research is about pattern recognition, image processing, and computer vision.
\end{IEEEbiography}

\begin{IEEEbiography}[{\includegraphics[width=1in,height=1.25in,clip,keepaspectratio]{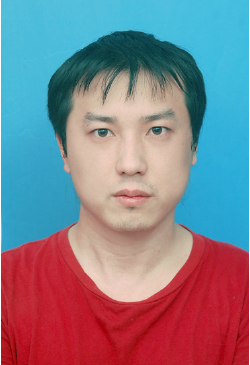}}]{Peng Zheng}
is a Ph.D. candidate at Hefei University of Technology since 2010. He received his Bachelor's degree in 2010 from Hefei University of Technology. His interests cover pattern recognition, image processing and computer vision.
\end{IEEEbiography}

\begin{IEEEbiography}[{\includegraphics[width=1in,height=1.25in,clip,keepaspectratio]{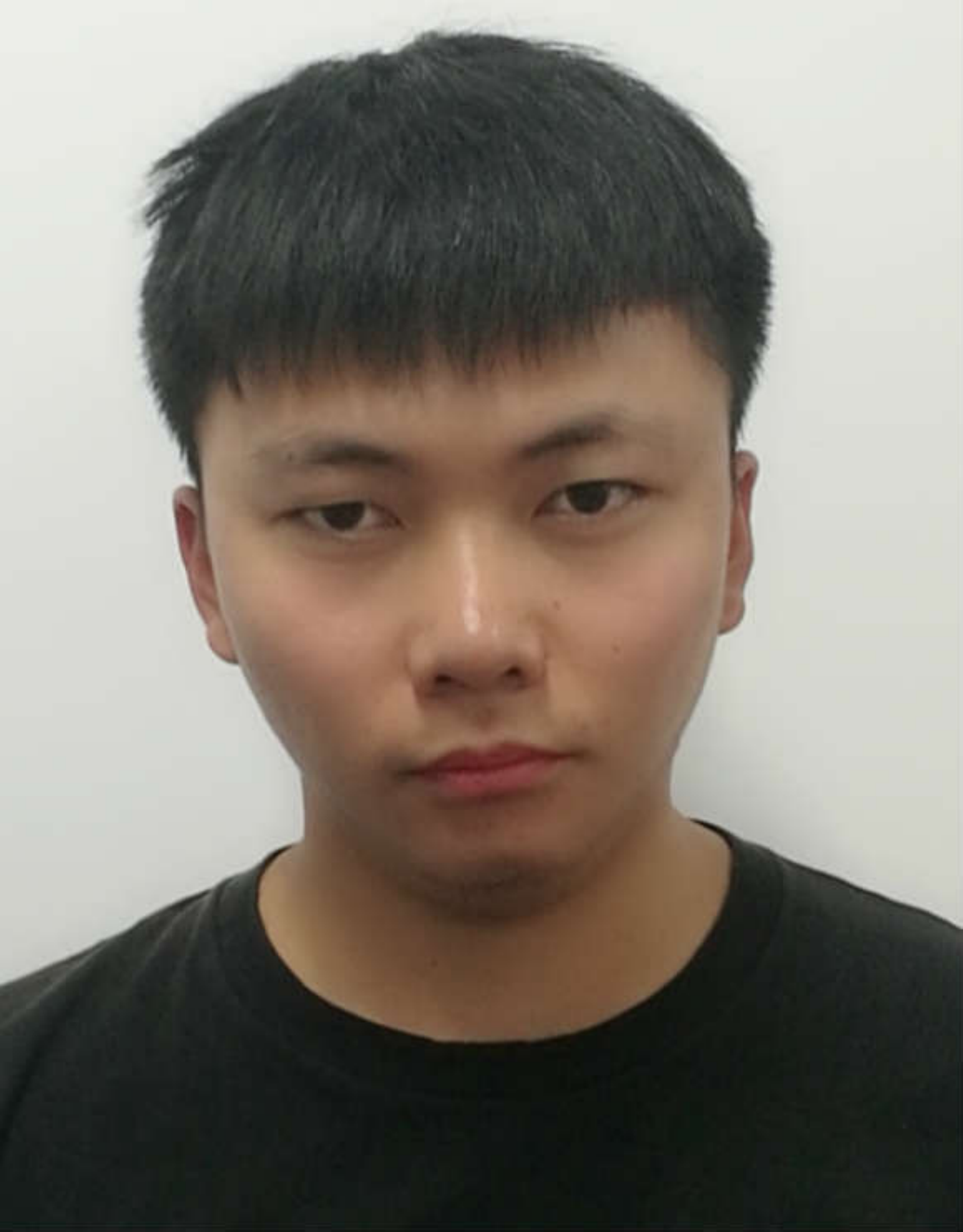}}]{Shou-tao Xu}
 is a Master student at Hefei University of Technology. His research interests cover pattern recognition, image processing, deep learning and computer vision.
\end{IEEEbiography}

\begin{IEEEbiography}[{\includegraphics[width=1in,height=1.25in,clip,keepaspectratio]{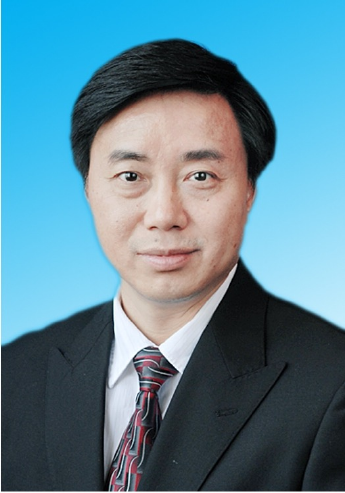}}]{Xindong Wu}
is an Alfred and Helen Lamson Endowed Professor in Computer Science, University of Louisiana at Lafayette (USA), and a Fellow of the IEEE and the AAAS. He received his Ph.D. degree in Artificial Intelligence from the University of Edinburgh, Britain. His research interests include data mining, knowledge-based systems, and Web information exploration. He is the Steering Committee Chair of the IEEE International Conference on Data Mining (ICDM), the Editor-in-Chief of Knowledge and Information Systems (KAIS, by Springer), and a Series Editor of the Springer Book Series on Advanced Information and Knowledge Processing (AI\&KP). He was the Editor-in-Chief of the IEEE Transactions on Knowledge and Data Engineering (TKDE, by the IEEE Computer Society) between 2005 and 2008.
\end{IEEEbiography}

%\newpage


%\vfill


\end{document}